\documentclass[10pt,twocolumn,letterpaper]{article}

\usepackage[pagenumbers]{cvpr} % To force page numbers, e.g. for an arXiv version

\usepackage{graphicx}
\usepackage{amsmath}
\usepackage{amssymb}
\usepackage{booktabs}
\usepackage{cases}
\usepackage{algorithm}
\usepackage{algpseudocode}
\usepackage{xcolor}
\usepackage{overpic}
\usepackage{rotating}
\usepackage[accsupp]{axessibility} % Improves PDF readability for those with disabilities.
\usepackage[pagebackref,breaklinks,colorlinks]{hyperref}

\usepackage[capitalize]{cleveref}
\crefname{section}{Sec.}{Secs.}
\Crefname{section}{Section}{Sections}
\Crefname{table}{Table}{Tables}
\crefname{table}{Tab.}{Tabs.}

\usepackage[pagebackref,breaklinks,colorlinks]{hyperref}

\usepackage[capitalize]{cleveref}
\crefname{section}{Sec.}{Secs.}
\Crefname{section}{Section}{Sections}
\Crefname{table}{Table}{Tables}
\crefname{table}{Tab.}{Tabs.}

\def\cvprPaperID{6035} % *** Enter the CVPR Paper ID here
\def\confName{CVPR}
\def\confYear{2023}

\begin{document}

\title{Denoising Diffusion Models for Plug-and-Play Image Restoration}

\author{Yuanzhi Zhu$^{1}$ \qquad Kai Zhang$^{1,}$\thanks{Corresponding author.} \qquad Jingyun Liang$^{1}$ \qquad Jiezhang Cao$^{1}$ \\Bihan Wen$^{2}$ \qquad Radu Timofte$^{1,3}$ \qquad Luc Van Gool$^{1,4}$\\% 
{$^{1}$ETH Z\"urich \quad $^{2}$Nanyang Technological University \quad $^{3}$University of W\"urzburg \quad $^{4}$KU Leuven}\\
{\tt\small yuazhu@student.ethz.ch \qquad kai.zhang@vision.ee.ethz.ch}
}

\maketitle

%%%%%%%%% ABSTRACT
\begin{abstract}

Plug-and-play Image Restoration (IR) has been widely recognized as a flexible and interpretable method for solving various inverse problems by utilizing any off-the-shelf denoiser as the implicit image prior. However, most existing methods focus on discriminative Gaussian denoisers. Although diffusion models have shown impressive performance for high-quality image synthesis, their potential to serve as a generative denoiser prior to the plug-and-play IR methods remains to be further explored.
While several other attempts have been made to adopt diffusion models for image restoration, they either fail to achieve satisfactory results or typically require an unacceptable number of Neural Function Evaluations (NFEs) during inference.
This paper proposes DiffPIR, which integrates the traditional plug-and-play method into the diffusion sampling framework. Compared to plug-and-play IR methods that rely on discriminative Gaussian denoisers, DiffPIR is expected to inherit the generative ability of diffusion models. Experimental results on three representative IR tasks, including super-resolution, image deblurring, and inpainting, demonstrate that DiffPIR achieves state-of-the-art performance on both the FFHQ and ImageNet datasets in terms of reconstruction faithfulness and perceptual quality with no more than 100 NFEs. The source code is available at {\url{https://github.com/yuanzhi-zhu/DiffPIR}}

\end{abstract}

%%%%%%%%% BODY TEXT
\section{Introduction}
\label{sec:intro}
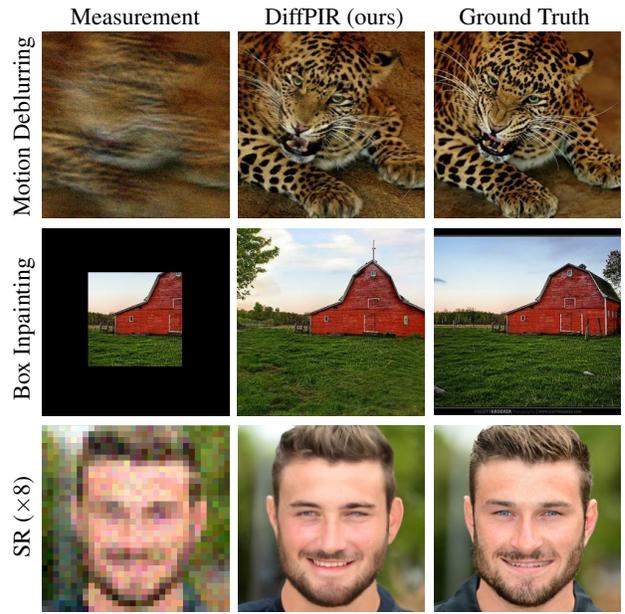
\begin{figure}
\centering
\begin{overpic}[width=1.0\linewidth]{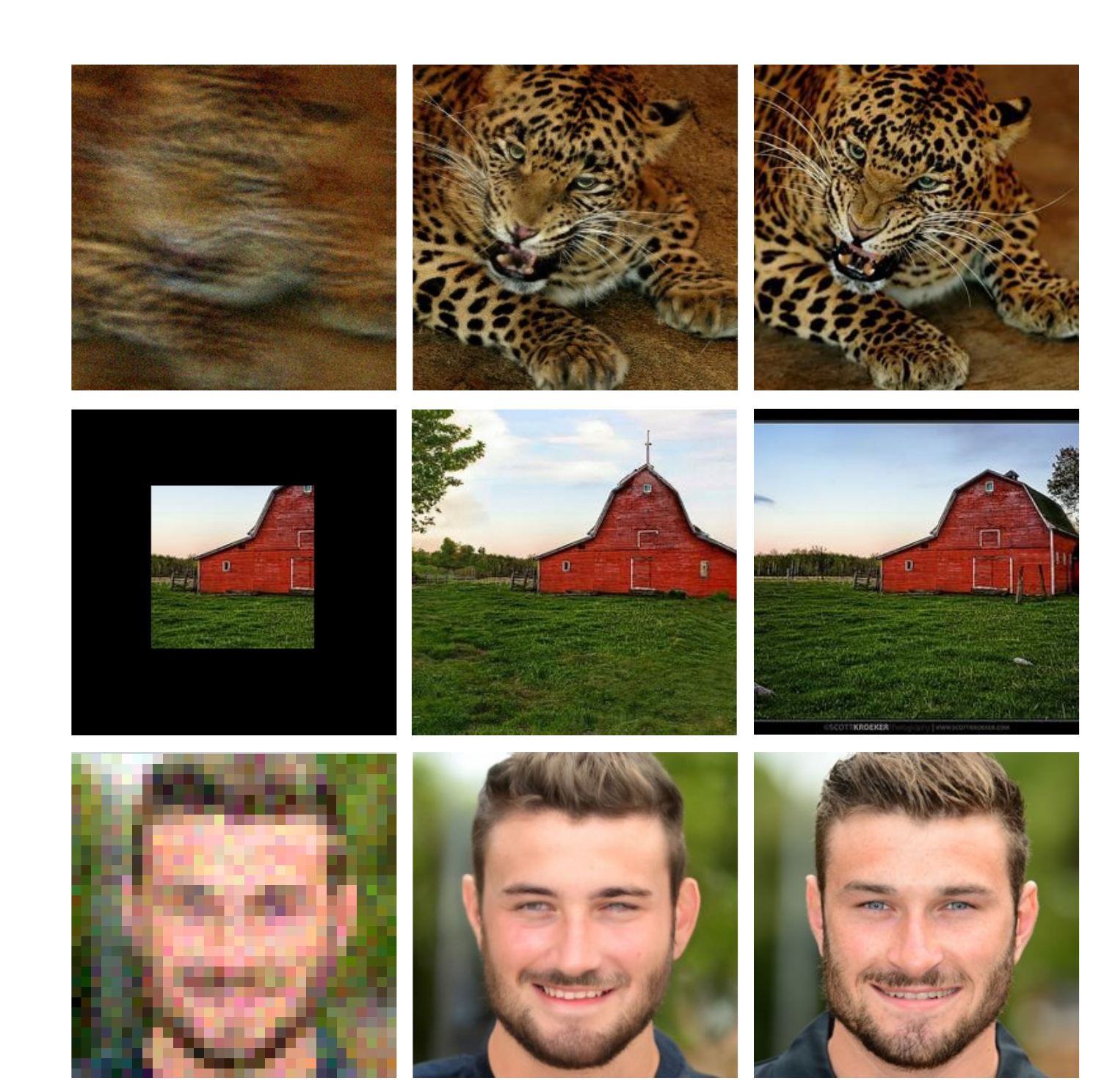}% no need to specify the file extension
\put(11,95){\color{black}{\small Measurement}}
\put(42,95){\color{black}{\small DiffPIR (ours)}}
\put(73,95){\color{black}{\small Ground Truth}}
\begin{turn}{90}
\put(10,-4.5){\color{black}{\small  SR ($\times$8)}}
\put(35,-4.5){\color{black}{\small Box Inpainting}}
\put(64.4,-4.5){\color{black}{\small Motion Deblurring}}
\end{turn}
\end{overpic}
\caption{\textbf{Restoration examples of DiffPIR.} We present the restored images and corresponding measurements and ground truth labels for several common image restoration tasks. 
}
\label{fig:intro}
\end{figure}

Recent studies have demonstrated that plug-and-play Image Restoration (IR) methods can effectively handle a variety of low-level vision tasks, such as image denoising~\cite{buades2005non}, image Super-Resolution (SR) \cite{dong2014learning,dong2015image,ledig2017photo}, image deblurring \cite{danielyan2011bm3d} and image inpainting \cite{iizuka2017globally}, with excellent results \cite{danielyan2010image,venkatakrishnan2013plug,chan2016plug,zhang2017learning,zhang2021plug}.

With the help of variable splitting algorithms, such as the Alternating Direction Method of Multipliers (ADMM) \cite{boyd2011distributed} and Half-Quadratic-Splitting (HQS) \cite{geman1995nonlinear}, plug-and-play IR methods integrate Gaussian denoisers into the iterative process, leading to improved performance and convergence.

The main idea of plug-and-play IR methods is to separate the data term and prior term of the following optimization problem
\begin{equation}\label{eq2}
  \hat{\mathbf{x}} = \mathop{\arg\min}_\mathbf{x}  \frac{1}{2\sigma_n^2}\|\mathbf{y} - \mathcal{H}(\mathbf{x})\|^2 + \lambda \mathcal{P}(\mathbf{x}),
\end{equation} 
where $\mathbf{y}$ is the measurement of ground truth $\mathbf{x}_0$ given the degradation model $\mathbf{y} = \mathcal{H}(\mathbf{x}_0) + \mathbf{n}$, $\mathcal{H}$ is a known degradation operator, $\sigma_n$ denotes the known standard deviation of i.i.d. Gaussian noise $\mathbf{n}$, and $\lambda\mathcal{P}(\cdot)$ is the prior term with regularization parameter $\lambda$.
To be specific, the data term ensures that the solution adheres to the degradation process, while the prior term enforces the solution follows the desired data distribution.
In particular, the prior term can be implicitly addressed by Gaussian denoisers \cite{meinhardt2017learning,zhang2019deep,venkatakrishnan2013plug,ryu2019plug}. 
Venkatakrishnan \etal \cite{venkatakrishnan2013plug} proposed to solve \eqref{eq2} by forming the augmented Lagrangian
function and using the ADMM technique with various image-denoising methods.
Kamilov \etal \cite{kamilov2017plug} used the BM3D denoising operator \cite{dabov2007image} to solve the prior subproblem for nonlinear inverse problems.
While the above methods used traditional denoisers, Zhang \etal \cite{zhang2017learning} made the first attempt to incorporate deep denoiser priors to solve various IR tasks. 
In subsequent research, Zhang \etal \cite{zhang2021plug} further proposed a more powerful denoiser for plug-and-play IR, which has since been adopted in numerous recent studies~\cite{bahnemiri2022learning,fermanian2022learned,kuo2022learning,alvar2022joint}.  
 
Unlike those traditional or convolutional neural network based {discriminative} Gaussian denoisers, denoisers parameterized by deep generative models are expected to better handle those ill-posed inverse problems due to their ability to model complex distributions.
Deep generative models such as Generative Adversarial Networks (GANs) \cite{goodfellow2020generative,karras2019style}, Normalizing Flows (NFs) \cite{dinh2016density} and Variational Autoencoders (VAEs) \cite{kingma2013auto,razavi2019generating} have been used as denoisers of plug-and-play IR framework \cite{dou2019pnp,kuznetsov2019prior,wei2022deep}. 
However, these generative models are not designed for denoising tasks and their generative capabilities are hindered when employed as plug-and-play prior.

Recently, diffusion models have demonstrated the ability to generate images with higher quality \cite{ramesh2022hierarchical,nichol2021improved,dhariwal2021diffusion} than previous generative models such as GANs, VAEs and NFs.
Diffusion models define a forward diffusion process that maps data to noise by gradually perturbing the input data with Gaussian noise.
While in the reverse process, they generate images by gradually removing Gaussian noise, with the intuition from non-equilibrium thermodynamics \cite{sohl2015deep}. 
The representative works in this area include Denoising Diffusion Probabilistic Models (DDPM) \cite{ho2020denoising} and score-based Stochastic Differential Equation (SDE) \cite{song2020score}. 
In addition to their unconditional generative power, diffusion models have also achieved remarkable success in the field of general inverse problems. Saharia \etal \cite{saharia2021image} employed a conditional network by using low-resolution images as conditional inputs to solve single-image SR. Lugmayr \etal \cite{lugmayr2022repaint} proposed an improved sampling strategy that resamples iterations for image inpainting. Chung \etal \cite{chung2022diffusion} introduced a Diffusion Posterior Sampling (DPS) method with Laplacian approximation for posterior sampling, which can be applied to noisy non-linear inverse problems. Choi \etal \cite{choi2021ilvr} proposed to adopt low-frequency information from measurement $\mathbf{y}$ to guide the generation process towards a narrow data manifold. Kawar \etal~\cite{kawar2022denoising} endorsed a time-efficient approach named Denoising Diffusion Restoration Models (DDRM) which performs diffusion sampling to reconstruct the missing information in $\mathbf{y}$ in the spectral space of $\mathcal{H}$ with Singular Value Decomposition (SVD). While the above methods achieve promising results, these methods either are hand-designed (\eg, \cite{lugmayr2022repaint}) or suffer from low sampling speed to get favorable performance (\eg, \cite{kawar2022denoising,chung2022diffusion}).

There exists another line of work, known as plug-and-play posterior sampling methods, which leverage the gradient of log posteriors to drive the samples to high-density regions. In this approach, the posteriors are decomposed into explicit likelihood functions and plug-and-play priors. Durmus \etal~\cite{durmus2018efficient} proposed a Moreau-Yosida regularised unadjusted Langevin algorithm for Bayesian computation such as inverse problems. Laumont \etal~\cite{laumont2022bayesian} extended this idea with Tweedie’s identity \cite{efron2011tweedie} and introduced PnP-unadjusted Langevin algorithm for image inverse problems. Both Romano \etal~\cite{romano2017little} and Kadkhodaie \etal~\cite{kadkhodaie2021stochastic} explicitly established the connection between a prior and a denoiser and used denoisers for stochastic posterior sampling. 

Inspired by the ability of plug-and-play IR to utilize any off-the-shelf denoisers as an implicit image prior, and considering that diffusion models are essentially generative denoisers, we propose denoising diffusion models for plug-and-play IR, referred to as DiffPIR. Following the plug-and-play IR method proposed in \cite{zhang2021plug}, we decouple the data term and the prior term and solve them iteratively within the diffusion sampling framework. The data term can be solved independently, allowing DiffPIR to handle a wide range of degradation models with various degradation operators $\mathcal{H}$. 
As for the prior term, it can be solved using an off-the-shelf diffusion model as a plug-and-play denoiser prior \cite{kawar2022denoising}.

We conduct experiments on different IR tasks such as SR, image deblurring and image inpainting on FFHQ \cite{karras2019style} and ImageNet \cite{russakovsky2015imagenet}.
By comparing our method with the other competitive approaches, we demonstrate that DiffPIR can efficiently restore images with superior quality (see the visual examples shown in Figure \ref{fig:intro}). 

%------------------------------------------------------------------------
\section{Background}
\label{sec:Background}

\subsection{Score-based Diffusion Models}
Diffusion is the process of destructing a signal (image) by adding Gaussian noise until the signal-to-noise ratio is negligible. This forward process can be described by an It\^{o} SDE \cite{song2020score}:
\begin{equation}\label{eq:itoSDE}
    \mathrm{d}\mathbf{x} = \mathbf{f}(\mathbf{x}, t) \mathrm{d}t + g(t) \mathrm{d} \mathbf{w},
\end{equation}
where $\mathbf{w}$ is the standard Wiener process, $\mathbf{f}(\cdot,t)$ is a vector-valued function called the drift coefficient, and ${g}(\cdot,t)$ is a scalar function known as the diffusion coefficient. 

The diffusion process described in \eqref{eq:itoSDE} can be reversed in time and has the form of \cite{anderson1982reverse}:
\begin{equation}\label{eq:revers_sde}
\mathrm{d}\mathbf{x} = [\mathbf{f}(\mathbf{x}, t) - g^2(t) \nabla_\mathbf{x} \log p_t(\mathbf{x})]\mathrm{d}t + g(t) \mathrm{d} \mathbf{w},
\end{equation}
where $p_t(\mathbf{x})$ is the marginal probability density at timestep $t$, and the only unknown part $\nabla_\mathbf{x} \log p_t(\mathbf{x})$ can be modelled as so-called score function $\mathbf{s}_\theta(\mathbf{x},t)$ with score matching methods \cite{hyvarinen2005estimation,song2019generative}. We utilize the convention of denoting the $\mathbf{x}$ at $t$ as $\mathbf{x}_t$ in subsequent discussions.

We can generate data samples according to \eqref{eq:revers_sde} by evaluating the score function $\mathbf{s}_\theta(\mathbf{x}_t,t)$ at each intermediate timestep during sampling, even if the initial state is Gaussian noise.
The training objective of time-dependent score function $\mathbf{s}_\theta(\mathbf{x}_t,t)$ with denoising score matching can be formulated as:
\begin{equation}\label{eq:objective}
\begin{aligned}
   \mathbb{E}_{t}\Big\{\gamma(t) \mathbb{E}_{\mathbf{x}_0}\mathbb{E}_{\mathbf{x}_t | \mathbf{x}_0 }
   \big[\|\mathbf{s}_\theta(\mathbf{x}_t, t) - \nabla_{\mathbf{x}_t}\log p_{0t}(\mathbf{x}_t | \mathbf{x}_0)\|_2^2 \big]\Big\},
\end{aligned}
\end{equation}
where $\gamma(t)$ is a positive weight coefficient, $t$ is uniformly sampled over $[0, T]$, $(\mathbf{x}_0,\mathbf{x}_t) \sim p_0(\mathbf{x}) p_{0t}(\mathbf{x}_t |\mathbf{x}_0)$. 
We can observe from \eqref{eq:objective} that a well-trained denoising score function $\mathbf{s}_\theta(\mathbf{x}_t, t)$ is also an ideal Gaussian denoiser under the circumstance that the transition probability $p_{0t}(\mathbf{x}_t | \mathbf{x}_0)$ is Gaussian.

\subsection{Denoising Diffusion Probabilistic Models}
For the specific choice of $\mathbf{f}(\mathbf{x},t)=-\frac{1}{2}\beta(t)\mathbf{x}$ and ${g}(\mathbf{x},t) = \sqrt{\beta(t)}$, we have the forward and reverse SDEs as the continuous version of the diffusion process in DDPM\cite{ho2020denoising}. One forward step of (discrete) DDPM is
\begin{equation}\label{eq:ddpm_forward}
\mathbf{x}_{t} = \sqrt{1-\beta_t} \mathbf{x}_{t-1} + \sqrt{\beta_t} \mathbf{\epsilon}_{t-1},
\end{equation}
with $\mathbf{\epsilon}_{t-1}\sim \mathcal{N}(\mathbf{0},\mathbf{I})$. 
The sample $\mathbf{x}_{t}$ is obtained by adding i.i.d. Gaussian noise with variance $\beta_t$ and scaling $\mathbf{x}_{t-1}$ with $\sqrt{1-\beta_t}$. In this way, the total variance is preserved and DDPM is also called ``Variance Preserving (VP)'' SDE \cite{song2020score}.
We can also sample $\mathbf{x}_{t}$ at an arbitrary timestep $t$ from $\mathbf{x}_{0}$ in closed form thanks to the good properties of Gaussian:
\begin{equation}\label{eq:ddpm_forward_arbitrary}
\mathbf{x}_{t} = \sqrt{\bar\alpha_t} \mathbf{x}_{0} + \sqrt{1-\bar\alpha_t} \mathbf{\epsilon},
\end{equation}
with new variance $1-\bar\alpha_t$ and scaling factor $\sqrt{\bar\alpha_t}$. 
In this work, $\{\beta_t\}$ is the noise schedule and we use the same notation $\alpha_t=1-\beta_t$ and $\bar\alpha_t = \prod_{s=1}^t \alpha_s$ following Ho \etal \cite{ho2020denoising}.
One reverse step of DDPM is
\begin{equation}\label{eq:ddpm_reverse}
\mathbf{x}_{t-1} = \frac{1}{\sqrt{\alpha_t}} \Big( \mathbf{x}_{t} - \frac{\beta_t}{\sqrt{1-\bar{\alpha}_t}} \mathbf{\epsilon}_\theta(\mathbf{x}_{t}, t) \Big) + \sqrt{\beta_t} \mathbf{\epsilon}_{t},
\end{equation}
where $\mathbf{\epsilon}_\theta(\mathbf{x}, t)$ is the function approximator intended to predict the total noise $\mathbf{\epsilon}$ between $\mathbf{x}_{t}$ and $\mathbf{x}_{0}$ in \eqref{eq:ddpm_forward_arbitrary}. 

In DDPM, the goal is to learn the noise added to $\mathbf{x}_{0}$; in score-based SDE, the goal is to learn the score function, the gradient of log-density of perturbed data; both with a U-Net. The connection between score function and noise prediction in DDPM can be formulated approximately as:
$\mathbf{s}_\theta(\mathbf{x}_{t}, t) \approx -\frac{\mathbf{\epsilon}_\theta(\mathbf{x}_t,t)}{\sqrt{1-\bar{\alpha}_t}}.$
From now on, we use both $\mathbf{\epsilon}_\theta(\mathbf{x},t)$ and $\mathbf{s}_\theta(\mathbf{x}, t)$ to represent diffusion models.

In order to sample with diffusion models more efficiently, Song \etal proposed Denoising Diffusion Implicit Models (DDIM) \cite{song2020denoising}, where the diffusion process can be extended from Markovian to non-Markovian and \eqref{eq:ddpm_reverse} can be rewritten as:
\begin{equation}\label{eq:ddim_reverse}
\begin{aligned}
\mathbf{x}_{t-1}=&\sqrt{\bar{\alpha}_{t-1}}\left(\frac{\mathbf{x}_{t}-\sqrt{1-\bar{\alpha}_{t}} \mathbf{\epsilon}_\theta(\mathbf{x}_{t}, t)}{\sqrt{\bar{\alpha}_{t}}}\right)\\
+&\sqrt{1-\bar{\alpha}_{t-1}-{\sigma_{{\eta}_t}}^{2}} \cdot \mathbf{\epsilon}_\theta(\mathbf{x}_{t}, t)+\sigma_{{\eta}_t} \mathbf{\epsilon}_{t},
\end{aligned}
\end{equation}
where $\mathbf{\epsilon}_{t}$ is standard Gaussian noise, the first term on the right-hand side is the predicted $\mathbf{x}_{0}$ at timestep $t$ scaled by $\sqrt{\bar{\alpha}_{t-1}}$, and the magnitude $\sigma_{{\eta}_t}$ of noise $\mathbf{\epsilon}_{t}$ controls how stochastic the diffusion process is.

\begin{figure*}[!t]
\vspace{-0.5cm}
\centering
\begin{overpic}[width=0.99\linewidth]{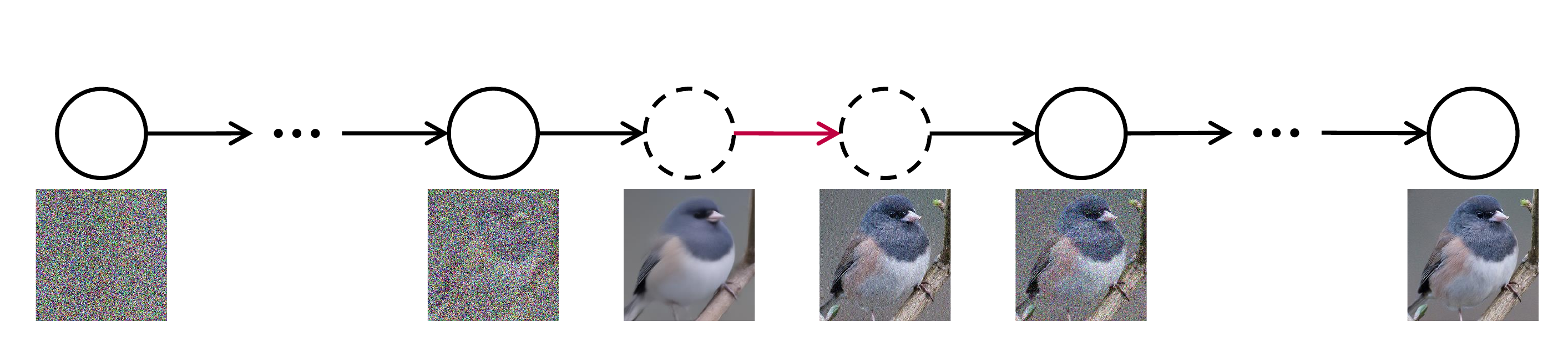}
% \put(33,19.5){\color{purple}{ $\mathop{\arg\min}_{\mathbf{x}} \|\mathbf{y} - \mathcal{H}(\mathbf{x})\|^2 + \rho_t\|\mathbf{x}-\mathbf{x}^{\scalebox{6}{$(t)$}}_0 \|^2$}}
\put(4.8,13.1){\color{black}{\scalebox{1.5}{$\mathbf{x}_T$}}}
\put(30.0,13.0){\color{black}{\scalebox{1.5}{$\mathbf{x}_t$}}}
\put(41.2,13.0){\color{black}{\scalebox{1.5}{ $\mathbf{x}^{\scalebox{0.6}{$(t)$}}_0$}}}
\put(53.8,13.0){\color{black}{\scalebox{1.5}{ $\mathbf{\hat{x}}^{\scalebox{0.6}{$(t)$}}_0$}}}
\put(65.9,13.0){\color{black}{\scalebox{1.5}{ $\mathbf{x}_{t\text{-}1}$}}}
\put(92.5,13.0){\color{black}{\scalebox{1.5}{$\mathbf{x}_0$}}}
%\end{sideways}
\end{overpic}
\vspace{-0.3cm}
\caption{\textbf{Illustration of our sampling method.}
For every state $\mathbf{x}_t$, following the prediction of the estimated $\mathbf{x}^{\scalebox{0.6}{$(t)$}}_0$ by the diffusion model, the measurement $\mathbf{y}$ is incorporated by solving the data proximal subproblem (indicated by the red arrow). Subsequently, the next state $\mathbf{x}_{t-1}$ is derived by adding noise back and thus completing one step of reverse diffusion sampling.
}
\label{fig:Demonstration}
\vspace{-0.3cm}
\end{figure*}
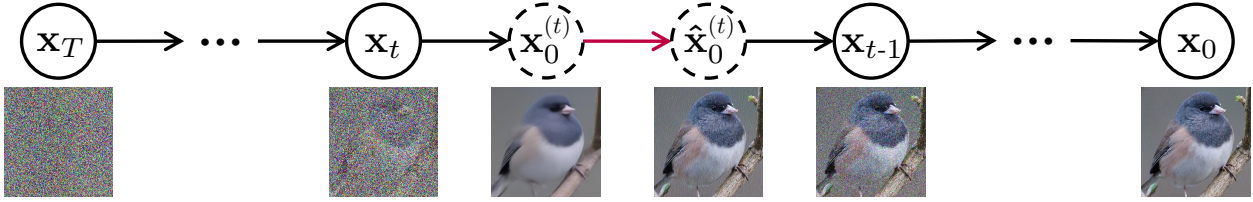

\subsection{Conditional Diffusion Models}
For conditional generation tasks given the condition $\mathbf{y}$, the goal is to sample images from the posterior distribution $p(\mathbf{x}|\mathbf{y})$.
In the work of Song \etal \cite{song2020score}, \eqref{eq:revers_sde} can be rewritten as follows for conditional generation with the help of Bayes' theorem
\begin{equation}\label{eq:cond_revers_sde}
\mathrm{d}\mathbf{x} = [\mathbf{f}(\mathbf{x}, t) - g^2(t) \nabla_\mathbf{x} (\log p_t(\mathbf{x}) + \log p_t(\mathbf{y}|\mathbf{x}))]\mathrm{d}t + g(t) \mathrm{d} \mathbf{w},
\end{equation}
where the posterior is divided into $p_t(\mathbf{x})$ and $p_t(\mathbf{y}|\mathbf{x})$. In this way, the unconditional pre-trained diffusion models can be used for conditional generation with an additional classifier. 

Ho \etal \cite{ho2022classifier} introduced the classifier-free diffusion guidance with $\mathbf{s}_\theta(\mathbf{x}, t,\mathbf{y})=\nabla_\mathbf{x}\log p_t(\mathbf{x}|\mathbf{y})$ the image-conditional diffusion models. With the same idea, Saharia \etal \cite{saharia2021image,saharia2022palette} trained image-conditional diffusion models for SR and image-to-image translation in concurrent work. Nichol \etal \cite{nichol2021glide} proposed to use text-guided diffusion models to generate photo-realistic images with classifier-free guidance. The hyperparameter $\lambda$ in \eqref{eq2} can be interpreted as the guidance scale in classifier-free diffusion models.

While the above methods need to train a diffusion model from scratch, conditional generation can also be done with unconditional pre-trained diffusion models \cite{choi2021ilvr,lugmayr2022repaint,chung2022diffusion,kawar2022denoising}. Given \eqref{eq:cond_revers_sde}, we can first update with one unconditional reverse diffusion step and then incorporate the conditional information.

%------------------------------------------------------------------------
\section{Proposed Method}
\label{sec:Method}

We adopt the HQS algorithm to decouple the data term and prior term in~\eqref{eq2}. This decoupling enables us to solve the decoupled subproblems iteratively and thus facilitates the utilization of diffusion sampling framework \cite{zhang2021plug}.
By introducing an auxiliary variable $\mathbf{z}$, \eqref{eq2} can be split into the following subproblems and be solved iteratively,
\begin{subequations}\label{eq:HQS}
\begin{numcases}{}
\mathbf{{z}}_{k}=\mathop{\arg\min}_{\mathbf{z}} \frac{1}{2(\sqrt{\lambda/\mu})^2}\|\mathbf{z}-\mathbf{x}_{k}\|^2  + \mathcal{P}(\mathbf{z})\label{eq:HQS_2}\\
\mathbf{{x}}_{k-1}=\mathop{\arg\min}_{\mathbf{x}}  \|\mathbf{y} - \mathcal{H}(\mathbf{x})\|^2 + \mu\sigma_n^2\|\mathbf{x}-\mathbf{{z}}_{k} \|^2 \label{eq:HQS_1},
\end{numcases}
\end{subequations}
where the parameter $\mu$ is introduced as the coefficient for the data-consistent constraint term.
Here the subproblem \eqref{eq:HQS_2} associated with prior term is a Gaussian denoising problem, and the subproblem \eqref{eq:HQS_1} associated with the data term is indeed a proximal operator \cite{parikh2014proximal} which usually has a closed-form solution. 

Our goal is to solve inverse problems via posterior sampling with generative diffusion models. 
Just like most plug-and-play methods, we can decouple the data term and prior term \cite{zhang2021plug}. 
The prior term ensures the generated sample is from the prior data distribution, and the data term narrows down the image manifold with the given measurement $\mathbf{y}$ \cite{choi2021ilvr}. We introduce them first in Section~\ref{sec:DiffusionDenoiser} and Section~\ref{sec:AnalyticSolution}, then our proposed sampling method in Section~\ref{sec:SamplingAlgo}. In Section~\ref{sec:Difference}, we highlight the differences between DiffPIR and several closely related diffusion-based methods. In Section~\ref{sec:AcceleratedSampling}, we show that our sampling can be accelerated like DDIM.

\subsection{Diffusion Models as Generative Denoiser Prior}
\label{sec:DiffusionDenoiser}
One important property of diffusion models is that the models can be understood as a combination of a generator (for the first few steps) and denoiser (for the rest of the steps) \cite{deja2022analyzing}. 
Intuitively, we can simply apply diffusion models as deep prior denoiser in HQS algorithm with a suitable initialization for plug-and-play IR \cite{zhang2021plug}. 
However, one significant difference between diffusion models and other deep denoisers is the generative power of diffusion models. With this generative ability, we will show that our method is capable of solving especially highly challenging inverse problems such as image inpainting with large masks.

It would be beneficial to build the connection between \eqref{eq:HQS_2} and the diffusion process. 
Assume we want to solve noiseless $\mathbf{z}_{k}$ from $\mathbf{x}_t$ with noise level $\bar{\sigma}_t=\sqrt{\frac{1-\bar\alpha_t}{\bar\alpha_t}}$, we can let $\sqrt{\lambda/\mu}=\bar{\sigma}_t$ for convenience. Given the noise schedule $\{\beta_t\}$ and hyperparameter $\lambda$, $\bar{\sigma}_t$ is known.
Indeed, \eqref{eq:HQS_2} can be solved as a {proximal} operator. 
Note that we have $\nabla_\mathbf{x} \mathcal{P}(\mathbf{x}) = -\nabla_\mathbf{x} \log p(\mathbf{x}) = -\mathbf{s}_\theta(\mathbf{x}) $, we can rewrite \eqref{eq:HQS_2} immediately as:
\begin{equation}\label{eq:HQS_2_sol1}
\mathbf{z}_{k} \approx \mathbf{{x}}_{k} + {\frac{1 - \bar\alpha_t}{\bar\alpha_t}} \mathbf{s}_\theta(\mathbf{{x}}_k),
\end{equation}
which means $\mathbf{z}_{k}$ is the estimated clear image $\mathbf{{x}}_{0}^{\scalebox{0.6}{$(t)$}}$ in ``Variance Exploding (VE)'' SDE form. Since the VP and VE Diffusion Models are actually equivalent to each other \cite{kawar2022denoising}, from now on we limit our discussion within DDPM without loss of generality. To make the discussion more clear, we rewrite \eqref{eq:HQS} as 
\begin{subequations}\label{eq:HQS_DMIR}
\begin{numcases}{}
\mathbf{{x}}_{0}^{\scalebox{0.6}{$(t)$}}=\mathop{\arg\min}_{\mathbf{z}} {\frac{1}{2\bar{\sigma}_t^2}\|\mathbf{z}-\mathbf{x}_{t}\|^2}  + {\mathcal{P}(\mathbf{z})}\label{eq:HQS_DMIR_2}\\
\mathbf{\hat{x}}_{0}^{\scalebox{0.6}{$(t)$}}=\mathop{\arg\min}_{\mathbf{x}}  \|\mathbf{y} - \mathcal{H}(\mathbf{x})\|^2 + \rho_t\|\mathbf{x}-\mathbf{{x}}_{0}^{\scalebox{0.6}{$(t)$}} \|^2 \label{eq:HQS_DMIR_1}\\
\mathbf{x}_{t-1} \longleftarrow \mathbf{\hat{x}}_{0}^{\scalebox{0.6}{$(t)$}}
\label{eq:HQS_DMIR_3},
\end{numcases}
\end{subequations}
where $\rho_t=\lambda (\sigma_n / \bar{\sigma}_t)^2$. Here \eqref{eq:HQS_DMIR_1} is the data subproblem to solve and \eqref{eq:HQS_DMIR_3} is a necessary step to finish our sampling method which will be introduced in \ref{sec:SamplingAlgo}.  
Additionally, we show in Appendix \ref{append:HQS_one_step} that we can also derive one-step reverse diffusion from HQS.

\begin{algorithm}[H]
    %\small
   \caption{DiffPIR}
   \label{alg:ddpir_gauss}
    \begin{algorithmic}[1]
     \Require $\mathbf{s}_\theta$, $T$, $\mathbf{y}$, $\sigma_{n}$,  ${\{\bar\sigma_{t}\}_{t=1}^T}$, $\zeta$, $\lambda$
    \State{Initialize $\mathbf{x}_{T}\sim \mathcal{N}(\mathbf{0}, \mathbf{I})$, pre-calculate $\rho_t \triangleq  {\lambda\sigma_{n}^2}/{\bar\sigma_{t}^2}$.}
      \For{$t=T$ {\bfseries to} $1$}
         \State{{$\mathbf{{x}}_{0}^{\scalebox{0.6}{$(t)$}} = \frac{1}{\sqrt{\bar\alpha_t}}(\mathbf{x}_t + (1 - \bar\alpha_t)\mathbf{s}_\theta(\mathbf{x}_t,t))$} \textcolor[rgb]{0.40,0.40,0.40}{\textit{// Predict $\mathbf{\hat{z}}_0$ with score model as denoisor}}}
        \State{\color{purple}{$ \mathbf{\hat{x}}_{0}^{\scalebox{0.6}{$(t)$}}=\mathop{\arg\min}_{\mathbf{x}} \|\mathbf{y} - \mathcal{H}(\mathbf{x})\|^2 + \rho_t\|\mathbf{x}-\mathbf{{x}}_{0}^{\scalebox{0.6}{$(t)$}} \|^2$} \textcolor[rgb]{0.40,0.40,0.40}{\textit{//  Solving data proximal subproblem}}}
         \State{$ \hat{\mathbf{\epsilon}} = \frac{1}{\sqrt{1 - \bar{\alpha}_t}} (\mathbf{x}_t - \sqrt{\bar{\alpha}_t}\mathbf{\hat{x}}_{0}^{\scalebox{0.6}{$(t)$}} )$ \textcolor[rgb]{0.40,0.40,0.40}{\textit{//  Calculate effective $\hat{\mathbf{\epsilon}}(\mathbf{x}_t,\mathbf{y})$}}}
         \State{$ \mathbf{\epsilon}_{t} \sim \mathcal{N}(\mathbf{0}, \mathbf{I})$}
         \State{$\mathbf{x}_{t-1}=\sqrt{\bar{\alpha}_{t-1}}\mathbf{\hat{x}}_{0}^{\scalebox{0.6}{$(t)$}}+ \sqrt{1 - \bar{\alpha}_{t-1}}(\sqrt{1-\zeta}\hat{\mathbf{\epsilon}} + \sqrt{\zeta}\mathbf{\epsilon}_{t})$  \textcolor[rgb]{0.40,0.40,0.40}{\textit{// Finish one step reverse diffusion sampling}}}
      \EndFor
      \State {\bfseries return} $\mathbf{x}_0$
    \end{algorithmic}
\end{algorithm}
\vspace{-0.3cm}

\begin{table*}[htbp]
\centering
\footnotesize
\resizebox{0.9\linewidth}{!}{% <------ Don't forget this %
\begin{tabular}{lcccccccccc}
\toprule
{\textbf{FFHQ}} & & \multicolumn{3}{c}{\textbf{Deblur (Gaussian)}} & \multicolumn{3}{c}{\textbf{Deblur (motion)}} & \multicolumn{3}{c}{\textbf{SR ($\times 4$)}}\\
\cmidrule(lr){3-5}
\cmidrule(lr){6-8}
\cmidrule(lr){9-11}
{\textbf{Method}}  & NFEs $\downarrow$& {PSNR $\uparrow$} & {FID $\downarrow$} & {LPIPS $\downarrow$} & {PSNR $\uparrow$} & {FID $\downarrow$} & {LPIPS $\downarrow$} & {PSNR $\uparrow$} & {FID $\downarrow$} & {LPIPS $\downarrow$}\\
\midrule
{DiffPIR~} & 100 & {27.36} & \textbf{59.65} & \textbf{0.236} & \textbf{26.57} &\textbf{65.78} & \textbf{0.255} & {26.64} & \textbf{65.77} & {0.260} \\
\cmidrule(l){1-11}
{DPS~\cite{chung2022diffusion}}  & 1000 & {25.46} &{65.57} & {0.247} & {23.31} &{73.31} & {0.289} & {25.77} & {67.01} & \textbf{0.256} \\
{DDRM~\cite{kawar2022denoising}} & 20 & 25.93 & 101.89 & 0.298 & - & - & -  & 27.92 & 89.43 & 0.265 \\
{DPIR~\cite{zhang2021plug}} & $>$20 & \textbf{27.79} & 123.99 & 0.450 & 26.41 & 146.44 & 0.467  & \textbf{28.03} & 133.39 & 0.456\\
\bottomrule
\end{tabular}}
\label{tab:results_noisy_ffhq}
\vspace{-0.2cm}
\end{table*}
\begin{table*}[htbp]
\centering
\resizebox{0.9\linewidth}{!}{% <------ Don't forget this %
\begin{tabular}{lcccccccccc}
\toprule
{\textbf{ImageNet}} & & \multicolumn{3}{c}{\textbf{Deblur (Gaussian)}} & \multicolumn{3}{c}{\textbf{Deblur (motion)}} & \multicolumn{3}{c}{\textbf{SR ($\times 4$)}}\\
\cmidrule(lr){3-5}
\cmidrule(lr){6-8}
\cmidrule(lr){9-11}
{\textbf{Method}}  & NFEs $\downarrow$ & {PSNR $\uparrow$} & {FID $\downarrow$} & {LPIPS $\downarrow$} & {PSNR $\uparrow$} & {FID $\downarrow$} & {LPIPS $\downarrow$} & {PSNR $\uparrow$} & {FID $\downarrow$} & {LPIPS $\downarrow$}\\
\midrule
{DiffPIR} & 100&  22.80 & \textbf{93.36}  & \textbf{0.355} & \textbf{24.01} & \textbf{124.63} & \textbf{0.366} &  23.18  &  \textbf{106.32}  & 0.371   \\
\cmidrule(l){1-11}
{DPS~\cite{chung2022diffusion}} & 1000 & 19.58 &138.80 &0.434 & 17.75 & 184.45 & 0.491 & 22.16  & 114.93  & 0.383  \\
{DDRM~\cite{kawar2022denoising}} & 20  & 22.33& 160.73 & 0.427 & - & - & -  & 23.89 & 118.55 & \textbf{0.358} \\
{DPIR~\cite{zhang2021plug}} & $>$20 & \textbf{23.86} & 189.92 & 0.476 & 23.60 & 210.31 & 0.489  & \textbf{23.99} & 204.83 & 0.475\\
\bottomrule
\end{tabular}
}
\caption{\textbf{Noisy quantitative results on FFHQ (top) and ImageNet (bottom).} 
We compute the average PSNR (dB), FID and LPIPS of different methods on Gaussian deblurring, motion deblurring and 4$\times$ SR.}
\label{tab:results_noisy_imagenet}
\vspace{-0.3cm}
\end{table*}

\subsection{Analytic Solution to Data Subproblem}
\label{sec:AnalyticSolution}
For IR tasks such as image deblurring, image inpainting and SR, we have a fast solution of \eqref{eq:HQS_DMIR_1} based on the estimated $\mathbf{z}_{0}$ on the image manifold \cite{zhang2021plug}. 
Since the data and prior terms are decoupled, the degradation model according to which we get the measurement $\mathbf{y}$ is only related to \eqref{eq:HQS_DMIR_1}. 

In cases where an analytical solution to \eqref{eq:HQS_DMIR_1} is not available, we can still approximate its solution using a first-order proximal operator method as follows:
\begin{equation}\label{eq:HQS_1_sol1}
\mathbf{\hat{x}}_{0}^{\scalebox{0.6}{$(t)$}} \approx \mathbf{{x}}_{0}^{\scalebox{0.6}{$(t)$}} - \frac{\bar{\sigma}_{t}^2}{2\lambda\sigma_{n}^2}\nabla_{\mathbf{{x}}_{0}^{\scalebox{0.6}{$(t)$}}} \|\mathbf{y} - \mathcal{H}(\mathbf{{x}}_{0}^{\scalebox{0.6}{$(t)$}})\|^2.
\end{equation}
This approximation can also be considered as one step of gradient descent, and we can calculate it numerically.

\begin{figure*}
\centering
\begin{overpic}[width=0.9\linewidth]{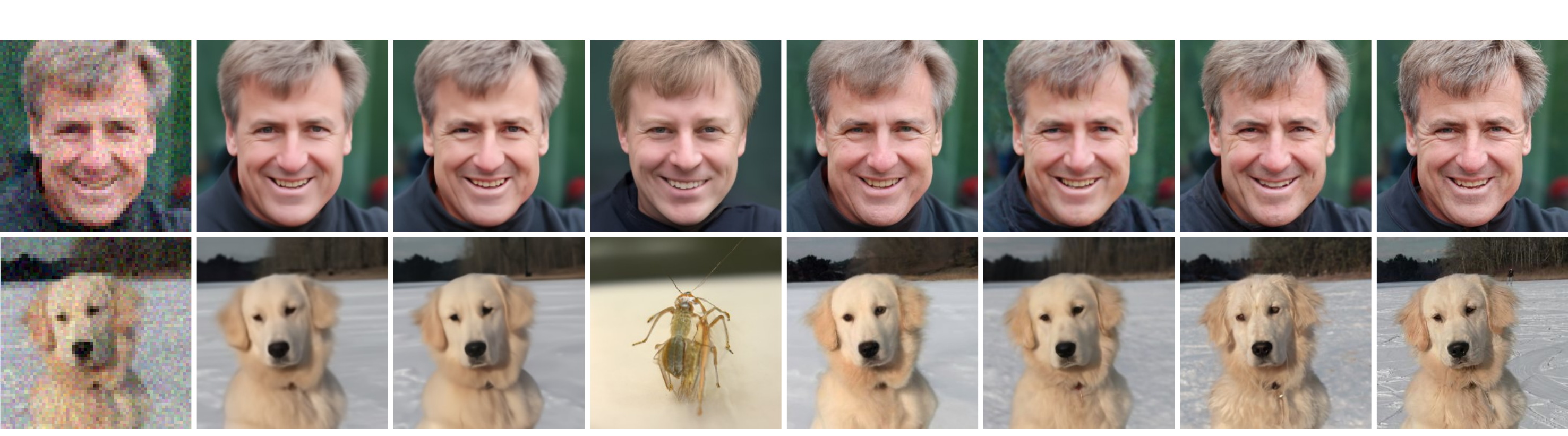}% no need to specify the file extension
\put(0.95,26){\color{black}{\small Measurement}}
\put(14.,26){\color{black}{\small DDRM (20)}}
\put(26.2,26){\color{black}{\small DDRM (100)}}
\put(39.8,26){\color{black}{\small DPS (100)}}
\put(52,26){\color{black}{\small DPS (1000)}}
\put(64,26){\color{black}{\small \textbf{DiffPIR} (20)}}
\put(75.8,26){\color{black}{\small \textbf{DiffPIR} (100)}}
\put(88.8,26){\color{black}{\small Ground Truth}}
%\end{sideways}
\end{overpic}
\caption{\textbf{Qualitative results of 4$\times$ SR.} We compare DiffPIR, DPS and DDRM with $\sigma_n=0.05$}
\label{fig:comparison_sr}
\vspace{-0.2cm}
\end{figure*}

\begin{figure*}
\centering
\begin{overpic}[width=0.9\linewidth]{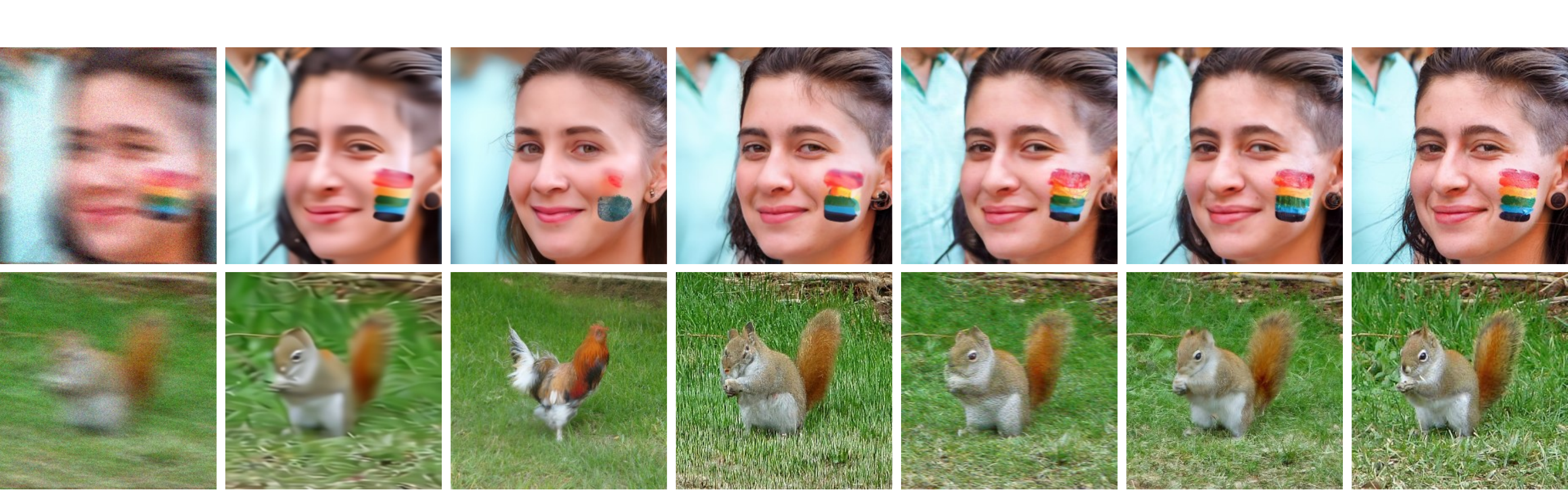}% no need to specify the file extension
\put(1.7,29.5){\color{black}{\small Measurement}}
\put(17.5,29.5){\color{black}{\small DPIR (20)}}
\put(32,29.5){\color{black}{\small DPS (100)}}
\put(46,29.5){\color{black}{\small DPS (1000)}}
\put(59.5,29.5){\color{black}{\small \textbf{DiffPIR} (20)}}
\put(73.4,29.5){\color{black}{\small \textbf{DiffPIR} (100)}}
\put(88,29.5){\color{black}{\small Ground Truth}}
\end{overpic}
\caption{\textbf{Qualitative results of motion deblurring.} We compare DiffPIR, DPS and DPIR with $\sigma_n=0.05$}
\label{fig:comparison_deblur}
\vspace{-0.2cm}
\end{figure*}
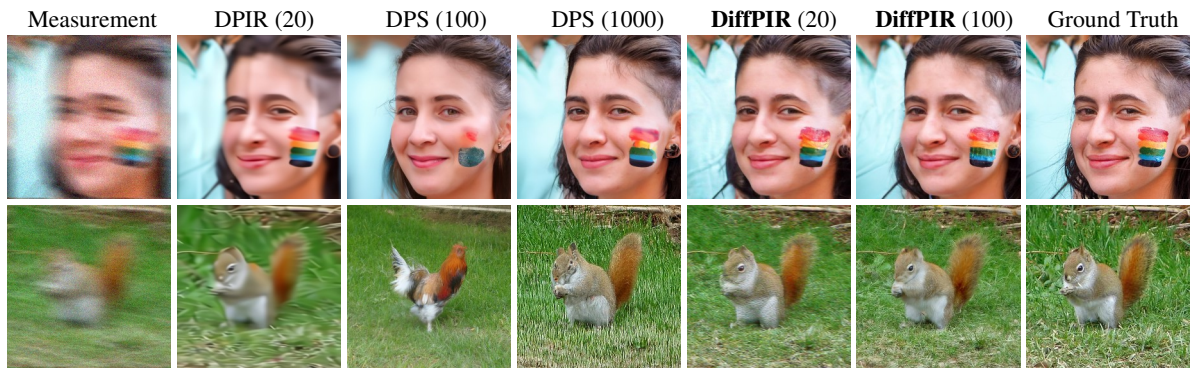

\subsection{DiffPIR Sampling}
\label{sec:SamplingAlgo}
With the discussion in the above two subsections, we can get an estimation of $\mathbf{\hat{x}}_{0}^{\scalebox{0.6}{$(t)$}} (\mathbf{y})$ given its noisy version $\mathbf{x}_{t}$ after calculating the analytical solution.
However, this estimation is not accurate, and we can add noise back to noise level $t-1$ as in \eqref{eq:HQS_DMIR_3}. % to finish one reverse diffusion step.
This estimation-correction idea was proposed in both \cite{song2020denoising} and \cite{karras2022elucidating}. 
In the DDIM fashion, we can first consider the estimation $\mathbf{\hat{x}}_{0}^{\scalebox{0.6}{$(t)$}} (\mathbf{y})$ as a sample from the conditional distribution $p(\mathbf{x}|\mathbf{y})$. 
Then we can calculate the effective predicted noise $ \hat{\mathbf{\epsilon}}(\mathbf{x}_t,\mathbf{y}) = \frac{1}{\sqrt{1 - \bar{\alpha}_t}} (\mathbf{x}_t - \sqrt{\bar{\alpha}_t}\mathbf{\hat{x}}_{0}^{\scalebox{0.6}{$(t)$}}  )$ to get the final one-step sampling expression similar to \eqref{eq:ddim_reverse}
\begin{equation}\label{eq:DDIMSampling}
    \mathbf{x}_{t-1}=\sqrt{\bar{\alpha}_{t-1}}\mathbf{\hat{x}}_{0}^{\scalebox{0.6}{$(t)$}} (\mathbf{y})+ \sqrt{1-\bar{\alpha}_{t-1}-\sigma_{{\eta}_t}^{2}} \hat{\mathbf{\epsilon}}(\mathbf{x}_t,\mathbf{y})+\sigma_{{\eta}_t} \mathbf{\epsilon}_{t},
\end{equation}
where $\hat{\mathbf{\epsilon}}$ is the corrected version of predicted noise and $\mathbf{\epsilon}_{t}\sim \mathcal{N}(\mathbf{0}, \mathbf{I})$.
% In our case, it's safe to set $\sigma_{{\eta}_t}=0$.
In our case, we found that the noise term $\sigma_{{\eta}_t} \mathbf{\epsilon}_{t}$ may not be strong enough and we can set $\sigma_{{\eta}_t}=0$.

Instead, we use a hyperparameter $\zeta$ to introduce noise to balance $\mathbf{\epsilon}_{t}$ and $\hat{\mathbf{\epsilon}}$ and the explicit form of \eqref{eq:HQS_DMIR_3} becomes
\begin{equation}\label{eq:DMIRSampling}
    \mathbf{x}_{t-1}=\sqrt{\bar{\alpha}_{t-1}}\mathbf{\hat{x}}_{0}^{\scalebox{0.6}{$(t)$}} + \sqrt{1 - \bar{\alpha}_{t-1}}(\sqrt{1-\zeta}\hat{\mathbf{\epsilon}} + \sqrt{\zeta}\mathbf{\epsilon}_{t}),
\end{equation}
 where the hyperparameter $\zeta$ controls the variance of the noise injected in each step and our sampling strategy becomes deterministic when $\zeta=0$.

Based on the above discussion, we summarized the detailed algorithm of our method, namely DiffPIR, in Algorithm \ref{alg:ddpir_gauss}. Our sampling method is demonstrated in Figure \ref{fig:Demonstration}. 
It is worth to mention that estimation of $\hat{x}_0^{\scalebox{0.6}{$(t)$}}(x_t,y)$ and the correction of $\hat\epsilon(x_t,y)$ involve the implicit computation of the conditional score $s_\theta(x_t,y)$.

\subsection{Comparison to Related Works}
\label{sec:Difference}
In this section, we will discuss the differences between the proposed DiffPIR and several closely related diffusion-based methods.

\vspace{0.2cm}
\noindent\textbf{DDRM \cite{song2020denoising}.} In DDRM, Kawar \etal introduced variational distribution of variables in the spectral space of general linear operator $\mathcal{H}$.
It is worth noting that DDRM is structurally similar to our method, as both first predict $\mathbf{x}_{0}$ and then add noise to forward sample $\mathbf{x}_{t-1}$.
However, DDRM can only work for linear operator $\mathcal{H}$, and its efficiency is not guaranteed when fast SVD is not feasible. On the contrary, DiffPIR can handle arbitrary degradation operator $\mathcal{H}$ with \eqref{eq:HQS_1_sol1}.

\vspace{0.2cm}
\noindent\textbf{DPS \cite{chung2022diffusion}.} In DPS, Chung \etal used Laplacian approximation to circumvent the intractability of posterior sampling, and their method can solve general noisy inverse problems. However, DPS suffers from its sampling speed and its reconstruction is not faithful with few sampling steps. Moreover, while DPS and DiffPIR have a similar solution for general inverse problems, just like the other posterior sampling methods with diffusion models (\ie, \cite{choi2021ilvr,lugmayr2022repaint,chung2022diffusion} and sampling methods in Appendix \ref{append:DPS_from_HQS}), it handles the measurement after each reverse diffusion step.
In contrast, DiffPIR adds measurement within reverse diffusion steps based on DDIM, which supports fast sampling.

%\vspace{-0.5cm}
\subsection{Accelerated Generation Process}
\label{sec:AcceleratedSampling}
While the generative ability of diffusion models has been proven to be better than other generative models like GAN and VAE, their slow inference speed ($\sim1000$ Neural Function Evaluations (NFEs)) impedes them from being applied in many real-world applications \cite{xiao2021tackling}. 
As suggested in \cite{song2020denoising}, DDPMs can be generalized to DDIMs with non-Markovian diffusion processes while still maintaining the same training objective. The underline reason is that the denoising objective \eqref{eq:objective} does not depend on any specific forward procedure as long as $p_{0t}(\mathbf{x}_t\mid\mathbf{x}_0)$ is fixed.  
As a result, our sampling sequence (length $T$) can be a subset of $[1,...,N]$ used in training.
To be specific, we adapt the quadratic sequence in DDIM, which has more sampling steps at low-noise regions and provides better reconstructions in our experiment \cite{song2020denoising}.

%------------------------------------------------------------------------
\section{Experiments}
\label{sec:Experiments}
\begin{table*}[htbp]
\centering
\resizebox{\linewidth}{!}{% <------ Don't forget this %
\begin{tabular}{lccccccccccccccc}
\toprule
{\textbf{FFHQ}} & & \multicolumn{2}{c}{\textbf{Inpaint (box)}} & \multicolumn{3}{c}{\textbf{Inpaint (random)}}&
\multicolumn{3}{c}{\textbf{Deblur (Gaussian)}} & \multicolumn{3}{c}{\textbf{Deblur (motion)}} & \multicolumn{3}{c}{\textbf{SR ($\times 4$)}} \\
\cmidrule(lr){3-4}
\cmidrule(lr){5-7}
\cmidrule(lr){8-10}
\cmidrule(lr){11-13}
\cmidrule(lr){14-16}
{\textbf{Method}} & NFEs $\downarrow$  & {FID $\downarrow$} & {LPIPS $\downarrow$} & {PSNR $\uparrow$} & {FID $\downarrow$} & {LPIPS $\downarrow$} & {PSNR $\uparrow$} & {FID $\downarrow$} & {LPIPS $\downarrow$} & {PSNR $\uparrow$} & {FID $\downarrow$} & {LPIPS $\downarrow$} & {PSNR $\uparrow$} & {FID $\downarrow$} & {LPIPS $\downarrow$}\\
\midrule
{DiffPIR} & 20 & {35.72} & {0.117} & { 34.03 } & { 30.81 } &{ 0.116 } & {30.74} & {46.64} & {0.170} & { 37.03 } &{20.11 } & {0.084 } & {29.17} & {58.02} & { 0.187 } \\
{DiffPIR} & 100 & \textbf{25.64} & \textbf{0.107} & \textbf{ 36.17 } & \textbf{ 13.68 } &\textbf{ 0.066 } & \textbf{31.00} & \textbf{39.27} & \textbf{0.152} & { 37.53 } &\textbf{11.54 } & \textbf{0.064 } & {29.52} & \textbf{47.80} & \textbf{ 0.174 } \\
\cmidrule(l){1-16}  
{DPS~\cite{chung2022diffusion}} & 1000 & {43.49} & {0.145} & {34.65 } & {33.14 } &{0.105} & {27.31 } &{51.23 } & {0.192 } & {26.73} &{58.63} & {0.222} & {27.64} & {59.06} & {0.209} \\
{DDRM~\cite{kawar2022denoising}} & 20 & 37.05  & 0.119 & 31.83 & 56.60 & 0.164 & 28.40  & 67.99  & 0.238  & - & - & -  & 30.09  & 68.59  & 0.188 \\
{DPIR~\cite{zhang2021plug}} & $>$20 & - & - & - & - & - & 30.52  & 96.16 & 0.350  &  \textbf{38.39} & 27.55  & 0.233  & \textbf{30.41}  & 96.16 & 0.362 \\
\bottomrule
\end{tabular}}
\caption{\textbf{Noiseless quantitative results on FFHQ.} We compute the average PSNR (dB), FID and LPIPS of different methods on inpainting, deblurring, and SR.}
\label{tab:results_noiseless_ffhq}
\vspace{-0.0cm}
\end{table*}
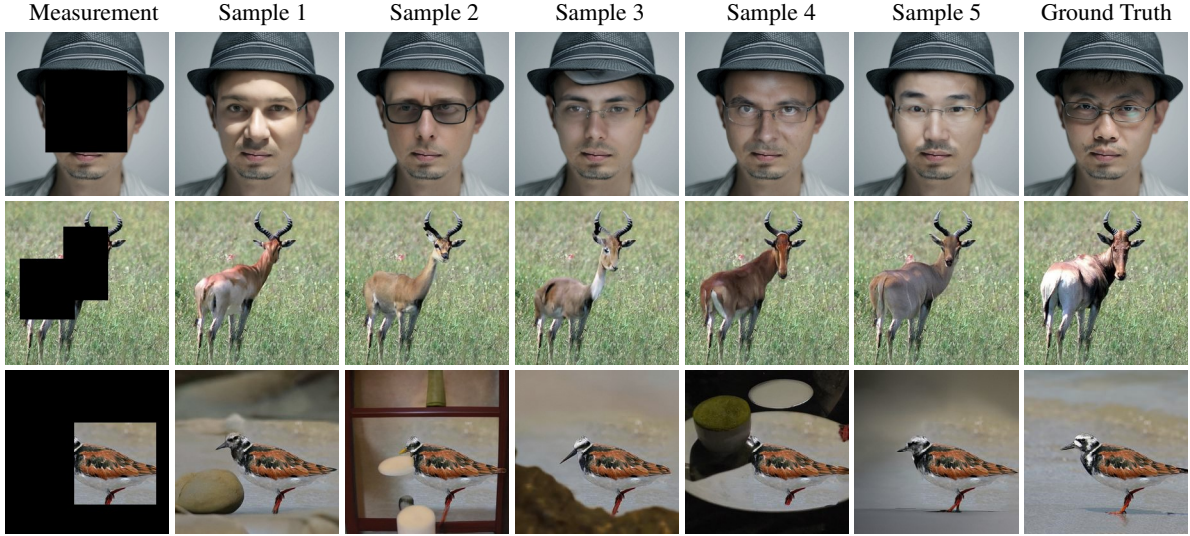
\begin{figure*}
\centering
\begin{overpic}[width=0.9\linewidth]{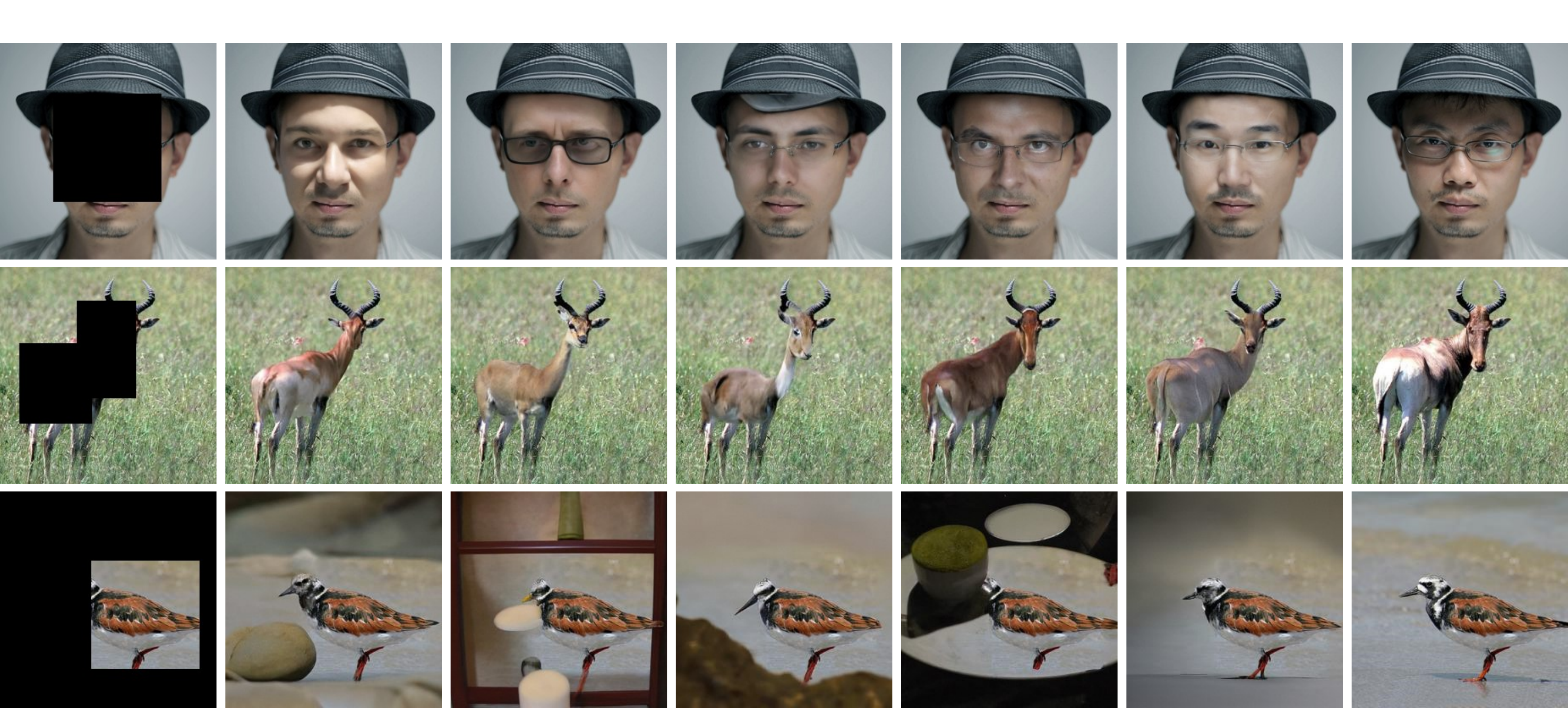}% no need to specify the file extension
\put(2,44){\color{black}{\small Measurement}}
\put(18,44){\color{black}{\small Sample 1}}
\put(32.5,44){\color{black}{\small Sample 2}}
\put(46.5,44){\color{black}{\small Sample 3}}
\put(61,44){\color{black}{\small Sample 4}}
\put(75,44){\color{black}{\small Sample 5}}
\put(87.6,44){\color{black}{\small Ground Truth}}
%\end{sideways}
\end{overpic}
\caption{\textbf{Qualitative results of inpainting.} We demonstrate the ability of DiffPIR to generate diverse reconstructions for different masks.}
\label{fig:inpainting}
\vspace{-0.2cm}
\end{figure*}

\subsection{Implementation Details}
We performed extensive experiments on the FFHQ 256$\times$256 \cite{karras2019style} and ImageNet 256$\times$256 \cite{russakovsky2015imagenet} datasets to evaluate different methods. For each dataset, we evaluate 100 hold-out validation images. As our method is training-free, we use pre-trained models from \cite{dhariwal2021diffusion} and \cite{choi2021ilvr} to conduct experiments on the ImageNet and FFHQ datasets, respectively. Throughout all experiments, we use the same linear noise schedule $\{\beta_t\}$ while employing different sampling sequences for each method.
Additionally, we keep all other settings of the diffusion models unchanged.

The degradation models are specified as follows: (\textit{i}) For inpainting with box-type mask, the mask is 128$\times$128 box region following the approach outlined in \cite{chung2022diffusion}; for inpainting with random-type mask, we mask out half of the total pixels at random; for inpainting with prepared mask images, we load the masks used in \cite{lugmayr2022repaint}. (\textit{ii}) When applying Gaussian blur, we utilize a blur kernel of size 61$\times$61 and a standard deviation of 3.0; for motion blur, the blur kernel is randomly generated with a size of 61$\times$61 and intensity value of 0.5 following the methodology described in \cite{chung2022diffusion}. To make a fair comparison, we use the same motion blur kernel for all experiments. 
(\textit{iii}) For SR, bicubic downsampling is performed. 
For image inpainting, we only consider the noiseless case. For image deblurring and SR, we do experiments with both noisy and noiseless settings. All images are normalized to the range of $[0, 1]$.
For additional experimental details, including parameter settings, please refer to Appendix \ref{append:Experimental_Details}.

\subsection{Quantitative Experiments}
For quantitative experiments, the metrics we used for comparison are Peak Signal-to-Noise Ratio (PSNR), Fr\'echet Inception Distance (FID), and Learned Perceptual Image Patch Similarity (LPIPS) distance. The FID evaluates the visual quality and distance between two image distributions. PSNR measures the faithfulness of restoration between two images, which is not important but necessary for IR tasks. LPIPS measures the perceptual similarity between two images. We report the results for both FFHQ 256$\times$256 and ImageNet 256$\times$256 datasets.

We compare DiffPIR (with 20 and 100 NFEs) with diffusion-based methods including DDRM \cite{kawar2022denoising} and DPS \cite{chung2022diffusion}, and plug-and-play method DPIR \cite{zhang2021plug}. The sampling steps for DDRM and DPS are 20 and 1000 according to the original paper. It is worth noting that since DPIR involves different iteration numbers for different tasks, we reported the minimum required number in the results. To ensure fairness, we employed the same pre-trained diffusion models and blur kernels for all methods in the comparison.

For noisy measurements with $\sigma_n=0.05$, we evaluate all methods on both datasets for 4$\times$ super-resolution (SR), Gaussian deblurring, and motion deblurring. However, we exclude DDRM from the motion deblurring evaluation since DDRM solely supports separable kernels for image deblurring.
Table \ref{tab:results_noisy_imagenet} demonstrates that DiffPIR achieves superior performance compared to all other comparison methods in terms of FID and LPIPS on both datasets. Additionally, DiffPIR achieves competitive scores in terms of PSNR. The only exception is observed in the LPIPS score for SR, which can be attributed to the potential introduction of accumulated errors during the sampling process due to the inaccuracy of the approximated bicubic kernels $\mathbf{k}$.

For noiseless measurement with $\sigma_n=0.0$,  we evaluate all methods on FFHQ 256$\times$256 for image inpainting, deblurring, and SR. DPIR is excluded from the inpainting experiments since it lacks initialization support for arbitrary masks.
The quantitative results are summarized in Table~\ref{tab:results_noiseless_ffhq}. For noiseless cases, our method with 100 NFEs outperforms the other comparison methods significantly in FID and LPIPS. Although DPIR exhibits a greater advantage in terms of PSNR for noiseless tasks, the generated images often fail to present high perceptual quality.
Remarkably, even with just 20 NFEs, DiffPIR showcases impressive competitive FID and LPIPS scores.

\subsection{Qualitative Experiments}
DiffPIR is able to produce high-quality reconstructions, as shown in Figure \ref{fig:intro} and Appendix \ref{append:VisualResults}. 
In Figure \ref{fig:comparison_sr}, we compare DiffPIR with DPS and DDRM on 4$\times$ SR. In Figure \ref{fig:comparison_deblur}, we compare DiffPIR with DPS and DPIR on motion deblurring. Our findings reveal that unlike DDRM and DPIR, which have a tendency to generate blurry images, DiffPIR excels in reconstructing images with intricate details. Moreover, compared to DPS, DiffPIR needs much fewer NFEs to get faithful reconstruction.

Furthermore, our sampling method has the ability to generate diverse reconstructions similar to DDPM.
With examples from image inpainting (see Figure \ref{fig:inpainting}) we show that DiffPIR can generate high-quality reconstruction with both diversity and good semantic alignment even when the degradation is strong (75\% masked).

\subsection{Ablation Study}
\label{sec:ablation}
\begin{figure}[!ht]
\vspace{-0.2cm}
\centering
\begin{overpic}[width=1.\linewidth]{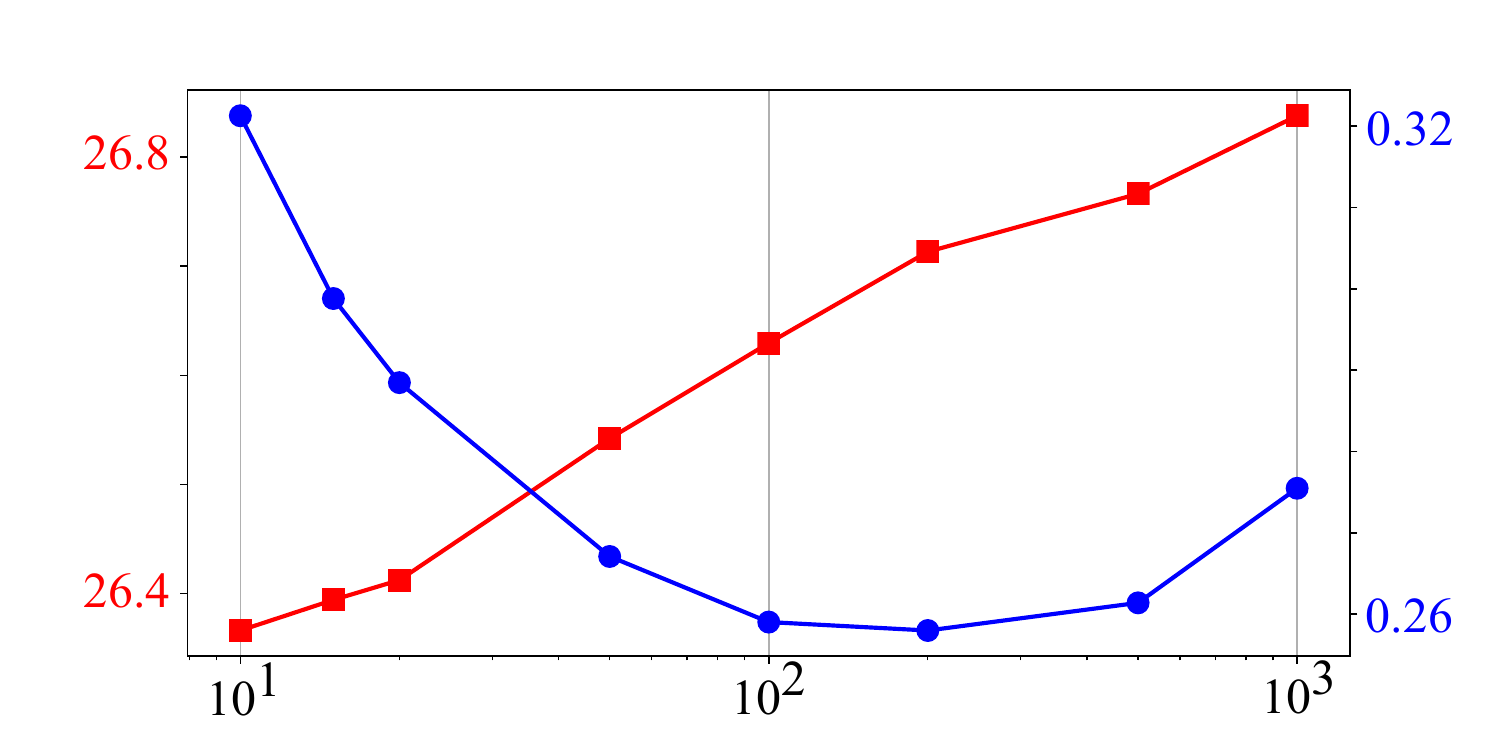}% no need to specify the file extension
\put(45,-1.5){\color{black}{\small NFEs}}
\begin{turn}{90}
\put(19,-5){\color{black}{\small \color{red}{PSNR/dB}}}
\put(22,-97){\color{black}{\small \color{blue}{LPIPS}}}
\end{turn}
\end{overpic}
\caption{Effect of sampling steps/NFEs}
\label{fig:ablation2}
\vspace{-0.2cm}
\end{figure}

\vspace{0.2cm}
\noindent\textbf{Effect of sampling steps.}
To investigate the effect of sampling steps or equivalently the number of NFEs, we perform 4$\times$ noisy SR ($\sigma_n{=}0.05$) experiment on 100 images from ImageNet validation set for sampling steps $T \in [10, 15, 20, 50, 100, 200, 500, 1000]$. Hyperparameters are fixed as $\lambda{=}8.0$ and $\zeta{=}0.3$, respectively.
 It is evident from Figure~\ref{fig:ablation2} that while the PSNR is log-linear to the number of NFEs, the LPIPS score is lowest for $T\in[100,500]$. Consequently, DiffPIR can produce detailed images with fewer than 100 NFEs and the default number of NFEs is set to 100 in this paper.

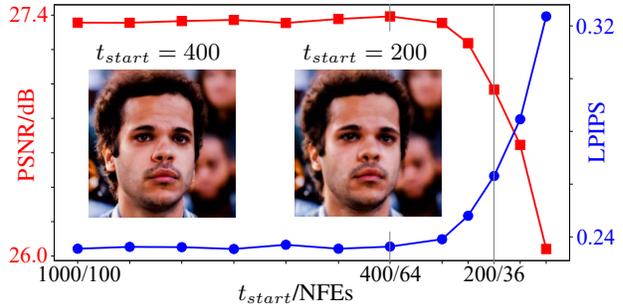
\begin{figure}
\centering
\begin{overpic}[width=1.\linewidth]{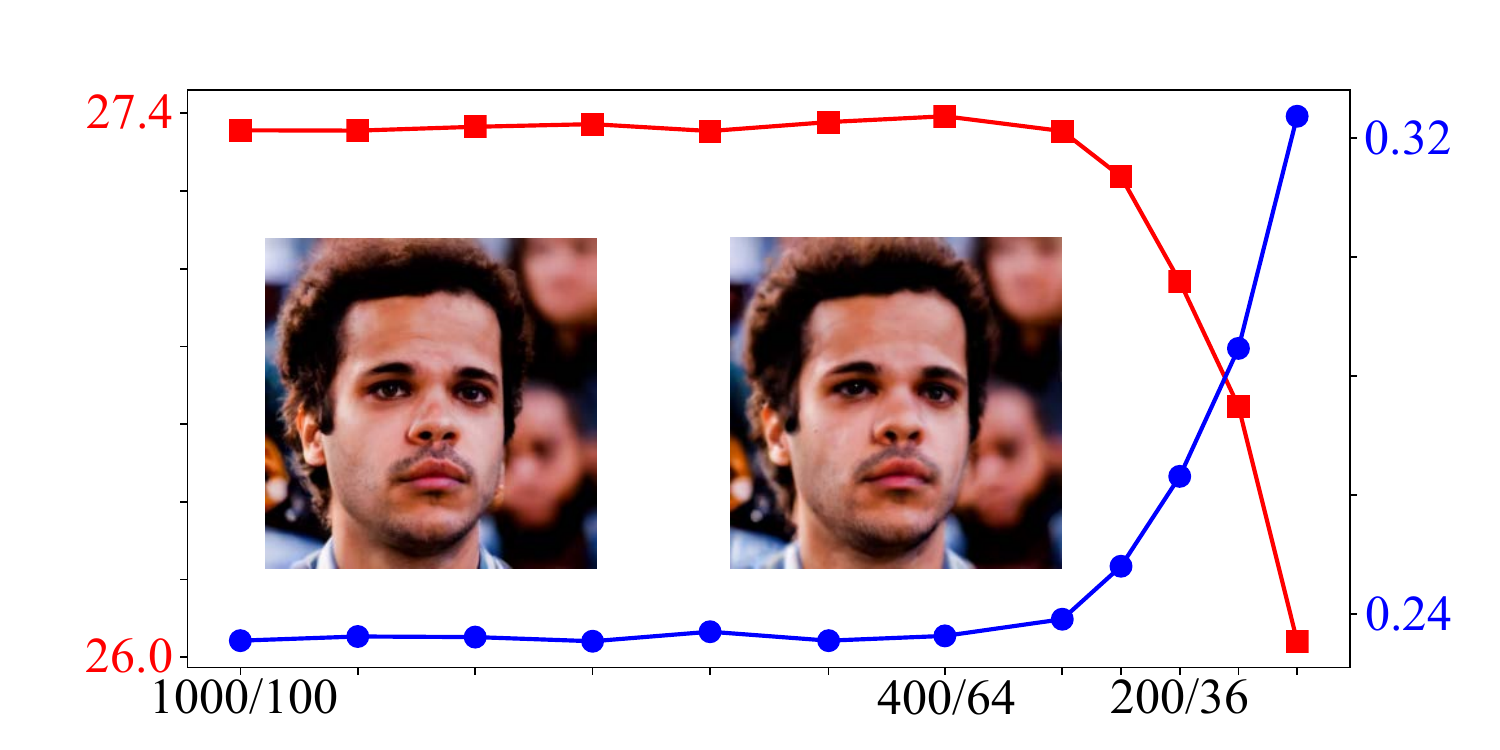}% no need to specify the file extension
\put(62.25,5){\color{gray}{\line(0,1){5}}}
\put(62.25,42){\color{gray}{\line(0,1){4.4}}}
\put(78.85,5){\color{gray}{\line(0,1){41.4}}}
\put(38,-1){\color{black}{\small $t_{start}$/NFEs}}
\put(15.4,37.5){\color{black}{\small $t_{start}=400$}}
\put(48.4,37.5){\color{black}{\small $t_{start}=200$}}
\begin{turn}{90}
\put(18,-5.5){\color{black}{\small \color{red}{PSNR/dB}}}
\put(21,-96.5){\color{black}{\small \color{blue}{LPIPS}}}
\end{turn}
\end{overpic}
\caption{Effect of $t_{start}$}
\label{fig:ablation_t_start}
\vspace{0.2cm}
\end{figure}

\vspace{0.2cm}
\noindent\textbf{Effect of $t_{start}$.} Similar to \cite{chung2022come}, our methods can also start the reverse diffusion process from a partial noised image rather than pure Gaussian noise to reduce the number of NFEs, especially for tasks like deblurring and SR. To analyze the impact of skipping the  initial diffusion steps on the ability to perform IR, we show in Figure \ref{fig:ablation_t_start} how PSNR and LPIPS change with the $t_{start}$ for noisy Gaussian deblurring task. 
Hyperparameters are fixed as $\lambda=8.0$ and $\zeta=0.5$. We find that our method performs well for even $t_{start}=400$, which leads to a great reduction of NFEs without loss of quality (see Appendix \ref{append:AblationStudy} for further explanation).
We also provide images for comparison with $t_{start}=200$ and $t_{start}=400$.

\begin{figure}
\vspace{-0.3cm}
\centering
\begin{overpic}[width=1\linewidth]{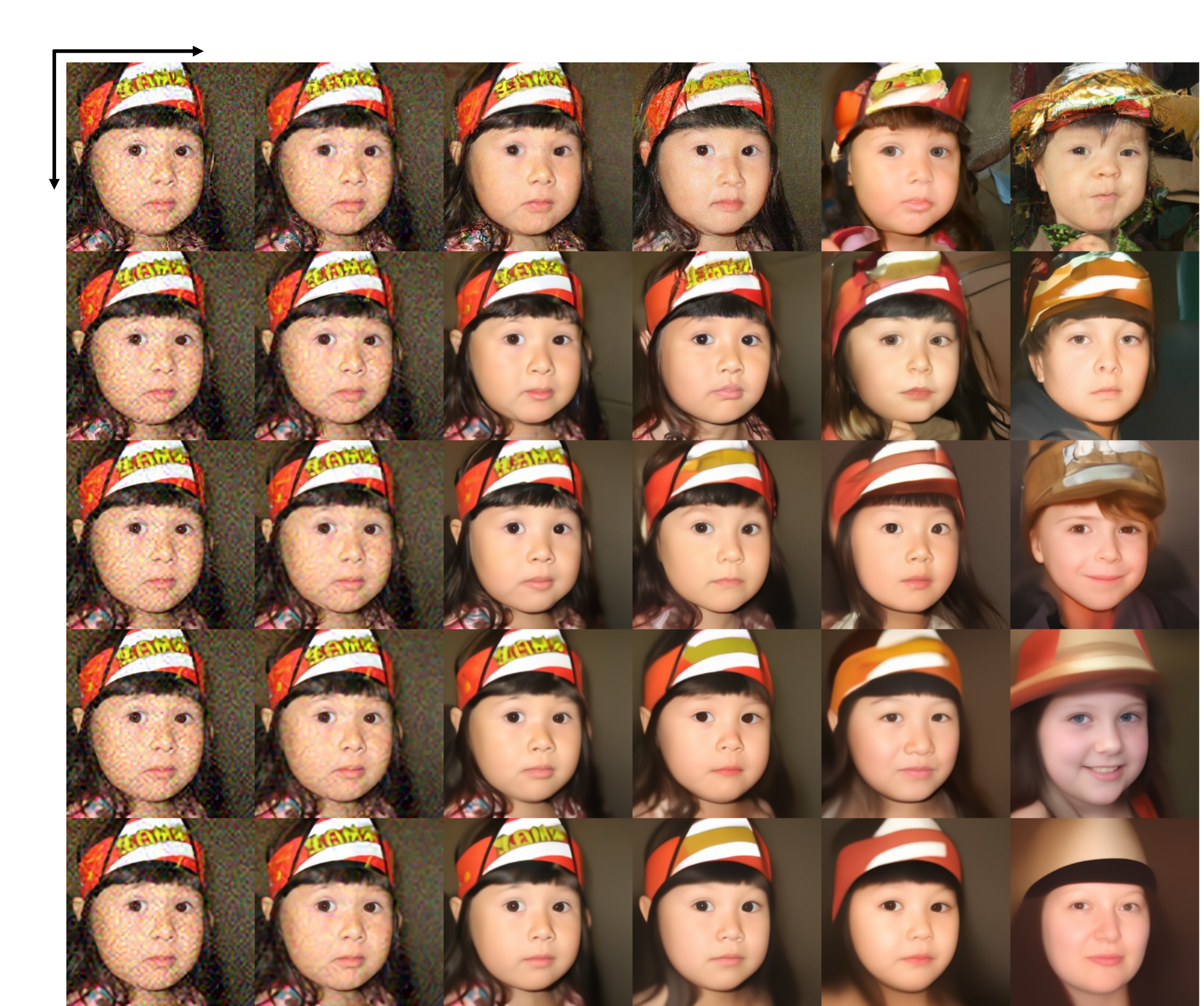}% no need to specify the file extension
\put(4,81){\color{black}{\small $\lambda$}}
\put(0.8,76){\color{black}{\small $\zeta$}}
\put(10,81){\color{black}{\small 0.1}}
\put(26,81){\color{black}{\small 1.0}}
\put(42,81){\color{black}{\small 10}}
\put(56,81){\color{black}{\small 100}}
\put(72,81){\color{black}{\small 1000}}
\put(86,81){\color{black}{\small 10000}}
\put(-2,7){\color{black}{\small 1.0}}
\put(-2,22){\color{black}{\small 0.8}}
\put(-2,38){\color{black}{\small 0.5}}
\put(-2,53){\color{black}{\small 0.2}}
\put(-2,68){\color{black}{\small 0.0}}
\end{overpic}
\caption{Effect of hyperparameters $\zeta$ and $\lambda$}
\label{fig:hyperp_para_ablation}
\vspace{-0.3cm}
\end{figure}
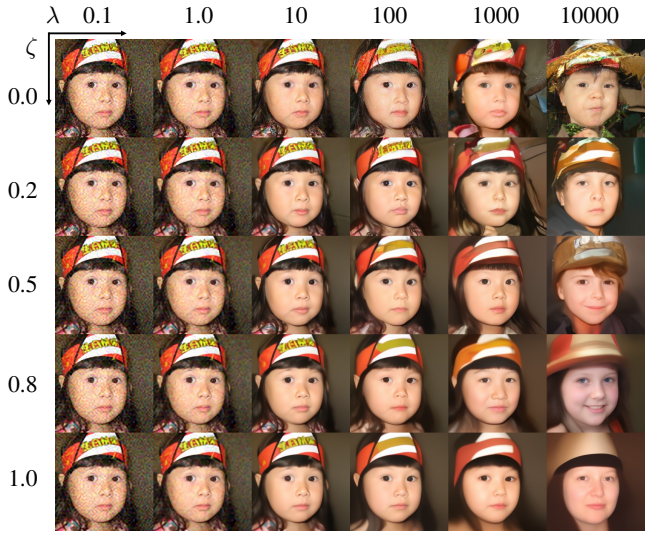

\vspace{0.2cm}
\noindent\textbf{Effects of $\lambda$ and $\zeta$.} DiffPIR has two hyperparameters $\lambda$ and $\zeta$, which control the strength of the condition guidance and the level of noise injected at each sampling timestep. 
To illustrate their effects, we show the reconstructed images of a motion-blurred sample in Figure~\ref{fig:hyperp_para_ablation}. 
Our observations from these results are as follows: (\textit{i}) when the guidance is too strong (\eg, $\lambda<1.0$) the noise is amplified whereas when the guidance is too weak (\eg, $\lambda>1000$), the generated images become more \textit{unconditional}; (\textit{ii}) the generated images tend to be blurry when $\zeta$ approaches 1. 

%------------------------------------------------------------------------
\section{Conclusions}
\label{sec:Conclusions}
In this paper, we introduce a new diffusion model-based sampling technique for plug-and-play image restoration, referred to as DiffPIR. Specifically, DiffPIR employs an HQS-based diffusion sampling approach that utilizes off-the-shelf diffusion models as plug-and-play denoising prior and solves the data subproblem in the clean image manifold.
Extensive experimental results highlight the superior flexibility, efficiency, and generalizability of DiffPIR in comparison to other competitive methods.

\vspace{1mm}
\noindent\textbf{Acknowledgements:}
This work was partly supported by the ETH Z\"urich General Fund (OK), the Alexander von Humboldt Foundation and the Huawei Fund.

% \clearpage
%%%%%%%%% REFERENCES
{\small
\bibliographystyle{ieee_fullname}
\bibliography{egbib}

\begin{thebibliography}{10}\itemsep=-1pt

\bibitem{alvar2022joint}
Saeed~Ranjbar Alvar, Mateen Ulhaq, Hyomin Choi, and Ivan~V Baji{\'c}.
\newblock Joint image compression and denoising via latent-space scalability.
\newblock {\em arXiv preprint arXiv:2205.01874}, 2022.

\bibitem{anderson1982reverse}
Brian~DO Anderson.
\newblock Reverse-time diffusion equation models.
\newblock {\em Stochastic Processes and their Applications}, 12(3):313--326,
  1982.

\bibitem{bahnemiri2022learning}
Sheyda~Ghanbaralizadeh Bahnemiri, Mykola Ponomarenko, and Karen Egiazarian.
\newblock Learning-based noise component map estimation for image denoising.
\newblock {\em IEEE Signal Processing Letters}, 29:1407--1411, 2022.

\bibitem{boyd2011distributed}
Stephen Boyd, Neal Parikh, Eric Chu, Borja Peleato, Jonathan Eckstein, et~al.
\newblock Distributed optimization and statistical learning via the alternating
  direction method of multipliers.
\newblock {\em Foundations and Trends{\textregistered} in Machine learning},
  3(1):1--122, 2011.

\bibitem{buades2005non}
Antoni Buades, Bartomeu Coll, and J-M Morel.
\newblock A non-local algorithm for image denoising.
\newblock In {\em 2005 IEEE computer society conference on computer vision and
  pattern recognition (CVPR'05)}, volume~2, pages 60--65. Ieee, 2005.

\bibitem{chan2016plug}
Stanley~H Chan, Xiran Wang, and Omar~A Elgendy.
\newblock Plug-and-play admm for image restoration: Fixed-point convergence and
  applications.
\newblock {\em IEEE Transactions on Computational Imaging}, 3(1):84--98, 2016.

\bibitem{choi2021ilvr}
Jooyoung Choi, Sungwon Kim, Yonghyun Jeong, Youngjune Gwon, and Sungroh Yoon.
\newblock Ilvr: Conditioning method for denoising diffusion probabilistic
  models.
\newblock {\em arXiv preprint arXiv:2108.02938}, 2021.

\bibitem{chung2022diffusion}
Hyungjin Chung, Jeongsol Kim, Michael~T Mccann, Marc~L Klasky, and Jong~Chul
  Ye.
\newblock Diffusion posterior sampling for general noisy inverse problems.
\newblock {\em arXiv preprint arXiv:2209.14687}, 2022.

\bibitem{chung2022come}
Hyungjin Chung, Byeongsu Sim, and Jong~Chul Ye.
\newblock Come-closer-diffuse-faster: Accelerating conditional diffusion models
  for inverse problems through stochastic contraction.
\newblock In {\em Proceedings of the IEEE/CVF Conference on Computer Vision and
  Pattern Recognition}, pages 12413--12422, 2022.

\bibitem{dabov2007image}
Kostadin Dabov, Alessandro Foi, Vladimir Katkovnik, and Karen Egiazarian.
\newblock Image denoising by sparse 3-d transform-domain collaborative
  filtering.
\newblock {\em IEEE Transactions on image processing}, 16(8):2080--2095, 2007.

\bibitem{danielyan2010image}
Aram Danielyan, Vladimir Katkovnik, and Karen Egiazarian.
\newblock Image deblurring by augmented lagrangian with bm3d frame prior.
\newblock In {\em Workshop on Information Theoretic Methods in Science and
  Engineering}, volume~1, 2010.

\bibitem{danielyan2011bm3d}
Aram Danielyan, Vladimir Katkovnik, and Karen Egiazarian.
\newblock Bm3d frames and variational image deblurring.
\newblock {\em IEEE Transactions on image processing}, 21(4):1715--1728, 2011.

\bibitem{deja2022analyzing}
Kamil Deja, Anna Kuzina, Tomasz Trzci{\'n}ski, and Jakub~M Tomczak.
\newblock On analyzing generative and denoising capabilities of diffusion-based
  deep generative models.
\newblock {\em arXiv preprint arXiv:2206.00070}, 2022.

\bibitem{dhariwal2021diffusion}
Prafulla Dhariwal and Alexander Nichol.
\newblock Diffusion models beat gans on image synthesis.
\newblock {\em Advances in Neural Information Processing Systems},
  34:8780--8794, 2021.

\bibitem{dinh2016density}
Laurent Dinh, Jascha Sohl-Dickstein, and Samy Bengio.
\newblock Density estimation using real nvp.
\newblock {\em arXiv preprint arXiv:1605.08803}, 2016.

\bibitem{dong2014learning}
Chao Dong, Chen~Change Loy, Kaiming He, and Xiaoou Tang.
\newblock Learning a deep convolutional network for image super-resolution.
\newblock In {\em European conference on computer vision}, pages 184--199.
  Springer, 2014.

\bibitem{dong2015image}
Chao Dong, Chen~Change Loy, Kaiming He, and Xiaoou Tang.
\newblock Image super-resolution using deep convolutional networks.
\newblock {\em IEEE transactions on pattern analysis and machine intelligence},
  38(2):295--307, 2015.

\bibitem{dou2019pnp}
Qi Dou, Cheng Ouyang, Cheng Chen, Hao Chen, Ben Glocker, Xiahai Zhuang, and
  Pheng-Ann Heng.
\newblock Pnp-adanet: Plug-and-play adversarial domain adaptation network at
  unpaired cross-modality cardiac segmentation.
\newblock {\em IEEE Access}, 7:99065--99076, 2019.

\bibitem{durmus2018efficient}
Alain Durmus, Eric Moulines, and Marcelo Pereyra.
\newblock Efficient bayesian computation by proximal markov chain monte carlo:
  when langevin meets moreau.
\newblock {\em SIAM Journal on Imaging Sciences}, 11(1):473--506, 2018.

\bibitem{efron2011tweedie}
Bradley Efron.
\newblock Tweedie’s formula and selection bias.
\newblock {\em Journal of the American Statistical Association},
  106(496):1602--1614, 2011.

\bibitem{fermanian2022learned}
Rita Fermanian, Mikael~Le Pendu, and Christine Guillemot.
\newblock Learned gradient of a regularizer for plug-and-play gradient descent.
\newblock {\em arXiv preprint arXiv:2204.13940}, 2022.

\bibitem{geman1995nonlinear}
Donald Geman and Chengda Yang.
\newblock Nonlinear image recovery with half-quadratic regularization.
\newblock {\em IEEE transactions on Image Processing}, 4(7):932--946, 1995.

\bibitem{goodfellow2020generative}
Ian Goodfellow, Jean Pouget-Abadie, Mehdi Mirza, Bing Xu, David Warde-Farley,
  Sherjil Ozair, Aaron Courville, and Yoshua Bengio.
\newblock Generative adversarial networks.
\newblock {\em Communications of the ACM}, 63(11):139--144, 2020.

\bibitem{ho2020denoising}
Jonathan Ho, Ajay Jain, and Pieter Abbeel.
\newblock Denoising diffusion probabilistic models.
\newblock {\em Advances in Neural Information Processing Systems},
  33:6840--6851, 2020.

\bibitem{ho2022classifier}
Jonathan Ho and Tim Salimans.
\newblock Classifier-free diffusion guidance.
\newblock {\em arXiv preprint arXiv:2207.12598}, 2022.

\bibitem{hyvarinen2005estimation}
Aapo Hyv{\"a}rinen and Peter Dayan.
\newblock Estimation of non-normalized statistical models by score matching.
\newblock {\em Journal of Machine Learning Research}, 6(4), 2005.

\bibitem{iizuka2017globally}
Satoshi Iizuka, Edgar Simo-Serra, and Hiroshi Ishikawa.
\newblock Globally and locally consistent image completion.
\newblock {\em ACM Transactions on Graphics (ToG)}, 36(4):1--14, 2017.

\bibitem{kadkhodaie2021stochastic}
Zahra Kadkhodaie and Eero Simoncelli.
\newblock Stochastic solutions for linear inverse problems using the prior
  implicit in a denoiser.
\newblock {\em Advances in Neural Information Processing Systems},
  34:13242--13254, 2021.

\bibitem{kamilov2017plug}
Ulugbek~S Kamilov, Hassan Mansour, and Brendt Wohlberg.
\newblock A plug-and-play priors approach for solving nonlinear imaging inverse
  problems.
\newblock {\em IEEE Signal Processing Letters}, 24(12):1872--1876, 2017.

\bibitem{karras2022elucidating}
Tero Karras, Miika Aittala, Timo Aila, and Samuli Laine.
\newblock Elucidating the design space of diffusion-based generative models.
\newblock {\em arXiv preprint arXiv:2206.00364}, 2022.

\bibitem{karras2019style}
Tero Karras, Samuli Laine, and Timo Aila.
\newblock A style-based generator architecture for generative adversarial
  networks.
\newblock In {\em Proceedings of the IEEE/CVF conference on computer vision and
  pattern recognition}, pages 4401--4410, 2019.

\bibitem{kawar2022denoising}
Bahjat Kawar, Michael Elad, Stefano Ermon, and Jiaming Song.
\newblock Denoising diffusion restoration models.
\newblock {\em arXiv preprint arXiv:2201.11793}, 2022.

\bibitem{kingma2013auto}
Diederik~P Kingma and Max Welling.
\newblock Auto-encoding variational bayes.
\newblock {\em arXiv preprint arXiv:1312.6114}, 2013.

\bibitem{kuo2022learning}
Pin-Hung Kuo, Jinshan Pan, Shao-Yi Chien, and Ming-Hsuan Yang.
\newblock Learning discriminative shrinkage deep networks for image
  deconvolution.
\newblock In {\em European Conference on Computer Vision}, pages 217--234.
  Springer, 2022.

\bibitem{kuznetsov2019prior}
Maxim Kuznetsov, Daniil Polykovskiy, Dmitry~P Vetrov, and Alex Zhebrak.
\newblock A prior of a googol gaussians: a tensor ring induced prior for
  generative models.
\newblock {\em Advances in Neural Information Processing Systems}, 32, 2019.

\bibitem{laumont2022bayesian}
R{\'e}mi Laumont, Valentin~De Bortoli, Andr{\'e}s Almansa, Julie Delon, Alain
  Durmus, and Marcelo Pereyra.
\newblock Bayesian imaging using plug \& play priors: when langevin meets
  tweedie.
\newblock {\em SIAM Journal on Imaging Sciences}, 15(2):701--737, 2022.

\bibitem{ledig2017photo}
Christian Ledig, Lucas Theis, Ferenc Husz{\'a}r, Jose Caballero, Andrew
  Cunningham, Alejandro Acosta, Andrew Aitken, Alykhan Tejani, Johannes Totz,
  Zehan Wang, et~al.
\newblock Photo-realistic single image super-resolution using a generative
  adversarial network.
\newblock In {\em Proceedings of the IEEE conference on computer vision and
  pattern recognition}, pages 4681--4690, 2017.

\bibitem{lugmayr2022repaint}
Andreas Lugmayr, Martin Danelljan, Andres Romero, Fisher Yu, Radu Timofte, and
  Luc Van~Gool.
\newblock Repaint: Inpainting using denoising diffusion probabilistic models.
\newblock In {\em Proceedings of the IEEE/CVF Conference on Computer Vision and
  Pattern Recognition}, pages 11461--11471, 2022.

\bibitem{meinhardt2017learning}
Tim Meinhardt, Michael Moller, Caner Hazirbas, and Daniel Cremers.
\newblock Learning proximal operators: Using denoising networks for
  regularizing inverse imaging problems.
\newblock In {\em Proceedings of the IEEE International Conference on Computer
  Vision}, pages 1781--1790, 2017.

\bibitem{nichol2021glide}
Alex Nichol, Prafulla Dhariwal, Aditya Ramesh, Pranav Shyam, Pamela Mishkin,
  Bob McGrew, Ilya Sutskever, and Mark Chen.
\newblock Glide: Towards photorealistic image generation and editing with
  text-guided diffusion models.
\newblock {\em arXiv preprint arXiv:2112.10741}, 2021.

\bibitem{nichol2021improved}
Alexander~Quinn Nichol and Prafulla Dhariwal.
\newblock Improved denoising diffusion probabilistic models.
\newblock In {\em International Conference on Machine Learning}, pages
  8162--8171. PMLR, 2021.

\bibitem{parikh2014proximal}
Neal Parikh, Stephen Boyd, et~al.
\newblock Proximal algorithms.
\newblock {\em Foundations and trends{\textregistered} in Optimization},
  1(3):127--239, 2014.

\bibitem{ramesh2022hierarchical}
Aditya Ramesh, Prafulla Dhariwal, Alex Nichol, Casey Chu, and Mark Chen.
\newblock Hierarchical text-conditional image generation with clip latents.
\newblock {\em arXiv preprint arXiv:2204.06125}, 2022.

\bibitem{razavi2019generating}
Ali Razavi, Aaron Van~den Oord, and Oriol Vinyals.
\newblock Generating diverse high-fidelity images with vq-vae-2.
\newblock {\em Advances in neural information processing systems}, 32, 2019.

\bibitem{romano2017little}
Yaniv Romano, Michael Elad, and Peyman Milanfar.
\newblock The little engine that could: Regularization by denoising (red).
\newblock {\em SIAM Journal on Imaging Sciences}, 10(4):1804--1844, 2017.

\bibitem{russakovsky2015imagenet}
Olga Russakovsky, Jia Deng, Hao Su, Jonathan Krause, Sanjeev Satheesh, Sean Ma,
  Zhiheng Huang, Andrej Karpathy, Aditya Khosla, Michael Bernstein, et~al.
\newblock Imagenet large scale visual recognition challenge.
\newblock {\em International journal of computer vision}, 115(3):211--252,
  2015.

\bibitem{ryu2019plug}
Ernest Ryu, Jialin Liu, Sicheng Wang, Xiaohan Chen, Zhangyang Wang, and Wotao
  Yin.
\newblock Plug-and-play methods provably converge with properly trained
  denoisers.
\newblock In {\em International Conference on Machine Learning}, pages
  5546--5557. PMLR, 2019.

\bibitem{saharia2022palette}
Chitwan Saharia, William Chan, Huiwen Chang, Chris Lee, Jonathan Ho, Tim
  Salimans, David Fleet, and Mohammad Norouzi.
\newblock Palette: Image-to-image diffusion models.
\newblock In {\em ACM SIGGRAPH 2022 Conference Proceedings}, pages 1--10, 2022.

\bibitem{saharia2021image}
Chitwan Saharia, Jonathan Ho, William Chan, Tim Salimans, David~J Fleet, and
  Mohammad Norouzi.
\newblock Image super-resolution via iterative refinement.
\newblock {\em arXiv preprint arXiv:2104.07636}, 2021.

\bibitem{sohl2015deep}
Jascha Sohl-Dickstein, Eric Weiss, Niru Maheswaranathan, and Surya Ganguli.
\newblock Deep unsupervised learning using nonequilibrium thermodynamics.
\newblock In {\em International Conference on Machine Learning}, pages
  2256--2265. PMLR, 2015.

\bibitem{song2020denoising}
Jiaming Song, Chenlin Meng, and Stefano Ermon.
\newblock Denoising diffusion implicit models.
\newblock {\em arXiv preprint arXiv:2010.02502}, 2020.

\bibitem{song2019generative}
Yang Song and Stefano Ermon.
\newblock Generative modeling by estimating gradients of the data distribution.
\newblock {\em Advances in Neural Information Processing Systems}, 32, 2019.

\bibitem{song2020score}
Yang Song, Jascha Sohl-Dickstein, Diederik~P Kingma, Abhishek Kumar, Stefano
  Ermon, and Ben Poole.
\newblock Score-based generative modeling through stochastic differential
  equations.
\newblock {\em arXiv preprint arXiv:2011.13456}, 2020.

\bibitem{venkatakrishnan2013plug}
Singanallur~V Venkatakrishnan, Charles~A Bouman, and Brendt Wohlberg.
\newblock Plug-and-play priors for model based reconstruction.
\newblock In {\em 2013 IEEE Global Conference on Signal and Information
  Processing}, pages 945--948. IEEE, 2013.

\bibitem{wei2022deep}
Xinyi Wei, Hans van Gorp, Lizeth Gonzalez-Carabarin, Daniel Freedman, Yonina~C
  Eldar, and Ruud~JG van Sloun.
\newblock Deep unfolding with normalizing flow priors for inverse problems.
\newblock {\em IEEE Transactions on Signal Processing}, 70:2962--2971, 2022.

\bibitem{xiao2021tackling}
Zhisheng Xiao, Karsten Kreis, and Arash Vahdat.
\newblock Tackling the generative learning trilemma with denoising diffusion
  gans.
\newblock {\em arXiv preprint arXiv:2112.07804}, 2021.

\bibitem{zhang2021plug}
Kai Zhang, Yawei Li, Wangmeng Zuo, Lei Zhang, Luc Van~Gool, and Radu Timofte.
\newblock Plug-and-play image restoration with deep denoiser prior.
\newblock {\em IEEE Transactions on Pattern Analysis and Machine Intelligence},
  2021.

\bibitem{zhang2017learning}
Kai Zhang, Wangmeng Zuo, Shuhang Gu, and Lei Zhang.
\newblock Learning deep cnn denoiser prior for image restoration.
\newblock In {\em Proceedings of the IEEE conference on computer vision and
  pattern recognition}, pages 3929--3938, 2017.

\bibitem{zhang2019deep}
Kai Zhang, Wangmeng Zuo, and Lei Zhang.
\newblock Deep plug-and-play super-resolution for arbitrary blur kernels.
\newblock In {\em Proceedings of the IEEE/CVF Conference on Computer Vision and
  Pattern Recognition}, pages 1671--1681, 2019.

\end{thebibliography}
}

\clearpage
%%%%%%%%% SUPPLEMENT
\setcounter{section}{0}
\renewcommand{\thesection}{\Alph{section}}

\section*{Appendix}

\section{Other HQS-based Sampling Methods}
\label{append:HQS_additional}

\subsection{HQS as one diffusion step}
\label{append:HQS_one_step}

For each of the conditional reverse diffusion step $t$, we are actually solving the MAP estimation problem on noise level $\beta_t$:
\begin{equation}\label{eq:constrained2}
  \begin{aligned}
  \hat{\mathbf{x}}_t &= \mathop{\arg\min}_{\mathbf{x}_t} ~ \frac{1}{2\sigma_n^2}\|\mathbf{y} - \mathcal{H}(\mathbf{x}_{t})\|^2 + \lambda \mathcal{P}(\mathbf{z}_{t})  \\ s.t. \quad \mathbf{x}_t&=\mathbf{z}_{t}= \sqrt{1-\beta_t}\mathbf{z}_{t-1}+\sqrt{\beta_t}\mathbf{\epsilon} 
  \end{aligned}
\end{equation}

With the HQS trick, now we have to solve 
\begin{subequations}\label{eq:HQSD}
\begin{numcases}{}
\mathbf{\hat{x}}_{t}=\mathop{\arg\min}_{\mathbf{x}_{t}} \|\mathbf{y} - \mathcal{H}(\mathbf{x}_{t})\|^2 + \mu\sigma_{n}^2\|\mathbf{x}_{t}-\mathbf{\hat{z}}_{t} \|^2 \label{eq:HQSD_1}\\
\mathbf{\hat{z}}_{t}=\mathop{\arg\min}_{\mathbf{z}_{t}} \frac{1}{2(\sqrt{\lambda/\mu})^2}\|\mathbf{z}_{t}-\mathbf{\hat{x}}_{t}\|^2  + \mathcal{P}(\mathbf{z}_{t})\label{eq:HQSD_2}
\end{numcases}
\end{subequations}
for each reverse diffusion step. We define $\sigma_{t}=\sqrt{\lambda/\mu}$ where $\sigma_{t}$ is the relative noise level between $\mathbf{x}_{t}$ and $\mathbf{z}_{t}$ with $\sigma_{t}=\sqrt{\frac{\beta_t}{1-\beta_t}}$.

To build the connection between \eqref{eq:HQSD_2} and a reverse diffusion step \eqref{eq:ddpm_reverse}, 
we first rewrite \eqref{eq:HQSD_2} as 
\begin{equation}\label{eq:HQSD_2p}
\begin{aligned}
\mathbf{\hat{z}}_{t-1}=&\mathop{\arg\min}_{\mathbf{z}_{t-1}} {\frac{1}{2({\frac{\beta_t}{1-\beta_t}})}\|\sqrt{1-\beta_t}\mathbf{z}_{t-1}+\sqrt{\beta_t}\mathbf{\epsilon} -\mathbf{\hat{x}}_{t}\|^2}  \\
+& \mathcal{P}(\sqrt{1-\beta_t}\mathbf{z}_{t-1}+\sqrt{\beta_t}\mathbf{\epsilon} ).
\end{aligned}
\end{equation}

Note that we have $\nabla_\mathbf{x} \mathcal{P}(\mathbf{x}) = -\nabla_\mathbf{x} \log p(\mathbf{x}) = -\mathbf{s}_\theta(\mathbf{x}) $. For any $\mathbf{\epsilon}_0$ sampled from $\mathcal{N}(\mathbf{0}, \mathbf{I})$, we have 
\begin{equation}\label{eq:HQSD_2_sol1}
\sqrt{1-\beta_t}\mathbf{\hat{z}}_{t-1}+\sqrt{\beta_t}\mathbf{\epsilon}_0 \approx \mathbf{\hat{x}}_{t} + \frac{\beta_t}{1-\beta_t}\mathbf{s}_\theta(\mathbf{\hat{x}}_t,t) 
\end{equation}
minimize the RHS of \eqref{eq:HQSD_2p} as first-order approximation of the proximal operator, which is also a standard gradient step with step length $\frac{\beta_t}{1-\beta_t}$.
Then $\mathbf{\hat{z}}_{t-1}$ can be solved as:
\begin{equation}\label{eq:HQSD_2_sol2}
\begin{aligned}
\mathbf{\hat{z}}_{t-1}=&\frac{1}{\sqrt{1-\beta_t}}\left(\mathbf{\hat{x}}_{t} + (\beta_t+{o}(\beta_t)) \mathbf{s}_\theta(\mathbf{\hat{x}}_t,t)  \right)\\
&+\sqrt{\beta_t}(1+{o}(\beta_t))\mathbf{\epsilon}'_0\\
\approx &\frac{1}{\sqrt{\alpha_t}}\left(\mathbf{\hat{x}}_{t} + \beta_t \mathbf{s}_\theta(\mathbf{\hat{x}}_t,t)  \right)+\sqrt{\beta_t} \mathbf{\epsilon}'_0
\end{aligned}
\end{equation}
where $\mathbf{\epsilon}'_0=-\mathbf{\epsilon}_0$ is also a sample from $\mathcal{N}(\mathbf{0}, \mathbf{I})$ and \eqref{eq:HQSD_2_sol2} is the same as reverse process of DDPM \eqref{eq:ddpm_reverse}. 
% Given the score function $\mathbf{s}_\theta(\mathbf{\hat{x}}_t,t)$, we can also evaluate $\mathbf{{z}}_0$ from the data distribution as
% \begin{equation}\label{eq:reverse_onestep}
% \mathbf{\hat{z}}_0 = \frac{1}{\sqrt{\bar\alpha_i}}(\mathbf{\hat{x}}_t + (1 - \bar\alpha_i)\mathbf{s}_\theta(\mathbf{\hat{x}}_t,i))
% \end{equation}
% with effective noise level $\bar{\sigma}_t=\sqrt{\frac{1 - \bar\alpha_i}{\bar\alpha_i}}$. This is equivalent to use the diffusion model as a denoiser as in \cite{zhang2021plug}. And then calculate $x_0$ with \eqref{eq:HQSD_1} at $t=0$.

\subsection{DPS as a Special Case}
\label{append:DPS_from_HQS}
For \eqref{eq:HQSD_1}, we can write similarly to Section~\ref{sec:AnalyticSolution}:
\begin{equation}\label{eq:HQSD_1_sol1}
\mathbf{\hat{x}}_{t} \approx \mathbf{\hat{z}}_{t} - \frac{{\sigma}_{t}^2}{2\lambda\sigma_{n}^2}\nabla_{\mathbf{z}_{t}} \|\mathbf{y} - \mathcal{H}(\mathbf{z}_{t})\|^2
\end{equation}
With the Theorem 1 from DPS\cite{chung2022diffusion}
\begin{equation}\label{eq:theorem1}
    \nabla_{\mathbf{{x}}_{t}} \log p_t(\mathbf{y}|\mathbf{{x}}_{t})
    \simeq  \nabla_{\mathbf{{x}}_{t}} \log p(\mathbf{y}|\mathbf{\hat{x}}_{0})
\end{equation}
\eqref{eq:HQSD_1_sol1} turned into:
\begin{equation}\label{eq:HQSD_1_sol12}
\mathbf{\hat{x}}_{t} \approx \mathbf{\hat{z}}_{t} - \frac{{\sigma}_{t}^2}{2\lambda\sigma_{n}^2}\nabla_{\mathbf{z}_{t}} \|\mathbf{y} - \mathcal{H}(\mathbf{z}_{0})\|^2
\end{equation}
By setting $\zeta_t=\frac{{\sigma}_{t}^2}{2\lambda\sigma_{n}^2}=\frac{1}{2\rho_{t}}$, we are now able to reproduce the sampling strategy in DPS.

Moreover, we can use the conclusion from \cite{song2020score} that
\begin{equation}\label{eq:theorem_score}
    \nabla_{\mathbf{{x}}_{t}} \log p_t({\mathbf{{x}}_{t}} \mid \mathbf{{y}}) \approx  \nabla_{\mathbf{{x}}_{t}} \log p_t({\mathbf{{x}}_{t}}) + \nabla_{\mathbf{{x}}_{t}} \log p_t({\mathbf{{y}}_{t}} \mid {\mathbf{{x}}_{t}}) \notag,
\end{equation}
where $\mathbf{y}_t=\sqrt{\bar{\alpha}_t}\mathbf{y}+ \sqrt{1-\bar{\alpha}_t}\epsilon$ is the measurement $\mathbf{y}$ at the given noise level and $\mathbf{y}_t$ is assumed to be the measurement from $\mathbf{x}_t$.

As a result, we can write a variant of \eqref{eq:HQSD_1_sol12} as
\begin{equation}\label{eq:HQSD_1_sol13}
\mathbf{\hat{x}}_{t} \approx \mathbf{\hat{z}}_{t} - \frac{{\sigma}_{t}^2}{2\lambda\sigma_{n}^2}\nabla_{\mathbf{z}_{t}} \|\mathbf{y}_{t} - \mathcal{H}(\mathbf{z}_{t})\|^2
\end{equation}
To distinguish them, we call the original DPS as DPS$y_0$ and the algorithm with \eqref{eq:HQSD_1_sol13} as DPS$y_t$.
The algorithm of DPS$y_t$ is:

\begin{algorithm}[H]
    \small
   \caption{Extended Sampling I: DPS$y_t$}
   \label{alg:ddpir_extend2}
    \begin{algorithmic}[1]
     \Require $\mathbf{s}_\theta$, $T$, $\mathbf{y}$, $\sigma_{n}$,  ${\{\sigma_{t}\}_{t=1}^T}$, $\lambda$
    \State{Initialize $\mathbf{x}_{T}\sim \mathcal{N}(\mathbf{0}, \mathbf{I})$}
      \For{$t=T$ {\bfseries to} $1$}
         \State{$ \mathbf{\epsilon}_{t} \sim \mathcal{N}(\mathbf{0}, \mathbf{I})$}
         \State{{$\mathbf{z}_{t-1} = \frac{1}{\sqrt{\alpha_t}} \Big( \mathbf{x}_{t} - \frac{\beta_t}{\sqrt{1-\bar{\alpha}_t}} \mathbf{\epsilon}_\theta(\mathbf{x}_{t}, t) \Big) + \sqrt{\beta_t} \mathbf{\epsilon}_{t}$} \textcolor[rgb]{0.40,0.40,0.40}{\textit{// one step reverse diffusion sampling}}}
        \State{\color{purple}{$ \mathbf{{x}}_{t-1}=\mathbf{{z}}_{t-1} - \frac{{\sigma}_{t}^2}{2\lambda\sigma_{n}^2}\nabla_{\mathbf{z}_{t-1}} \|\mathbf{y}_{t-1} - \mathcal{H}(\mathbf{z}_{t-1})\|^2$} \textcolor[rgb]{0.40,0.40,0.40}{\textit{//  Solving data proximal subproblem}}}
      \EndFor
      \State {\bfseries return} $\mathbf{x}_0$
    \end{algorithmic}
\end{algorithm}

\section{Experimental Details}
\label{append:Experimental_Details}
    
\subsection{Hyperparameters Values}
We list the hyperparametrs values for different tasks and datasets in \ref{tab:Hyperparameters2}.

\begin{table}[htbp]
\centering
\resizebox{\linewidth}{!}{% <------ Don't forget this %
\begin{tabular}{lcccccc}
\toprule
{\textbf{NFE=20}} & \multicolumn{4}{c}{\textbf{$\sigma_y=0.05$}} & \multicolumn{2}{c}{\textbf{$\sigma_y=0.0$}} \\
\cmidrule(lr){2-5}
\cmidrule(lr){6-7}
Dataset & \multicolumn{2}{c}{\textbf{FFHQ 256x256}} & \multicolumn{2}{c}{\textbf{ImgaeNet 256x256}}  & \multicolumn{2}{c}{\textbf{FFHQ 256x256}}\\
\cmidrule(lr){2-3}
\cmidrule(lr){4-5}
\cmidrule(lr){6-7}
Hyperparameters & {$\lambda$} & {$\zeta$} & {$\lambda$} & {$\zeta$} & {$\lambda$} & {$\zeta$}\\
\midrule
\textbf{Inpaint (box)}  & - & - & - & -  & 6.0 & 1.0 \\
\textbf{Inpaint (random)}  & - & - & - & -  & 3.0 & 1.0 \\
\textbf{Deblur (gauss)}  & 8.0 & 0.5  & 12.0 & 0.9 & 15.0 & 0.5 \\
\textbf{Deblur (motion)}  & 7.0 & 0.8 & 7.0 & 1.0  & 25.0 & 1.0 \\
\textbf{SR ($\times 4$)}  & 8.0 & 0.4 & 10.0 & 0.5  & 9.0 & 0.2 \\
\bottomrule
\end{tabular}
}
% \caption{Hyperparameters for different tasks.}
\label{tab:Hyperparameters1}
\vspace{-0.3cm}
\end{table}
\begin{table}[htbp]
\centering
\resizebox{\linewidth}{!}{% <------ Don't forget this %
\begin{tabular}{lcccccc}
\toprule
{\textbf{NFE=100}} & \multicolumn{4}{c}{\textbf{$\sigma_y=0.05$}} & \multicolumn{2}{c}{\textbf{$\sigma_y=0.0$}} \\
\cmidrule(lr){2-5}
\cmidrule(lr){6-7}
Dataset & \multicolumn{2}{c}{\textbf{FFHQ 256x256}} & \multicolumn{2}{c}{\textbf{ImgaeNet 256x256}}  & \multicolumn{2}{c}{\textbf{FFHQ 256x256}}\\
\cmidrule(lr){2-3}
\cmidrule(lr){4-5}
\cmidrule(lr){6-7}
Hyperparameters & {$\lambda$} & {$\zeta$} & {$\lambda$} & {$\zeta$} & {$\lambda$} & {$\zeta$}\\
\midrule
\textbf{Inpaint (box)}  & - & - & - & -  & 6.0 & 0.5 \\
\textbf{Inpaint (random)}  & - & - & - & -  & 7.0 & 1.0 \\
\textbf{Deblur (gauss)}  & 7.0 & 0.3  & 8.0 & 0.3 & 12.0 & 0.4 \\
\textbf{Deblur (motion)}  & 7.0 & 0.4 & 8.0 & 0.7 & 7.0 & 0.9 \\
\textbf{SR ($\times 4$)}  & 8.0& 0.2 & 9.0 & 0.5  & 6.0 & 0.3 \\
\bottomrule
\end{tabular}
}
\caption{Hyperparameters for different tasks.}
\label{tab:Hyperparameters2}
\vspace{-0.3cm}
\end{table}
    
\subsection{Closed-form Solutions}
In this section, we will introduce the specific degradation models and fast solutions of \eqref{eq:HQS_DMIR_1} for image restoration tasks including SR, deblurring and inpainting.

\vspace{0.2cm}
\noindent  \textbf{Image Inpainting.} In this work, we only consider the noiseless inpainting.
The degradation model of masked image for inpainting can be expressed as
\begin{equation}\label{eq:inpainting}
  \mathbf{y} = \mathbf{M}\odot\mathbf{x},
\end{equation}
where $\mathbf{M}$ is any user-defined mask and is a matrix with boolean elements, and $\odot$ denotes element-wise multiplication. The image inpainting task is to recover the missing pixels from the known pixels as $\mathbf{y}$.
The closed-from solution of \eqref{eq:HQS_DMIR_1} is given by \cite{zhang2021plug}
\begin{equation}\label{eq_dm}
  \mathbf{x}_{0} = \frac{\mathbf{M}\odot\mathbf{y} + \rho_t\mathbf{z}_{0}}{\mathbf{M} + \rho_t},
\end{equation}
and the division here is also element-wise.

\vspace{0.2cm}
\noindent  \textbf{Image Deblurring.} The linear degradation model for image deblurring with Gaussian noise is generally expressed as
\begin{equation}\label{eq:deblur}
  \mathbf{y} = \mathbf{x}\otimes \mathbf{k} + \mathbf{n},
\end{equation}
where $\otimes$ is two-dimensional convolution operator applied on all image channels.
%Note that we consider the uniform image deblurring which is also termed as image deconvolution.
By assuming $\otimes$ is also a circular convolution operator, the analytical solution of \eqref{eq:HQS_DMIR_1} is given by \cite{zhang2021plug}
\begin{equation}\label{eq:deblur_closedform}
  \mathbf{x}_{0} = \mathcal{F}^{-1}\left(\frac{\overline{\mathcal{F}(\mathbf{k})}\mathcal{F}(\mathbf{y})+\rho_t\mathcal{F}(\mathbf{z}_{0}) }{\overline{\mathcal{F}(\mathbf{k})} \mathcal{F}(\mathbf{k})+\rho_t}\right),
\end{equation}
where the $\mathcal{F}(\cdot)$ and $\mathcal{F}^{-1}(\cdot)$ denote Fast Fourier Transform (FFT) and its inverse.

\vspace{0.2cm}
\noindent  \textbf{Single Image Super-Resolution (SISR).} In this work, we consider bicubic SR, which has the following degradation model
\begin{equation}\label{eq:sisr_bicubic_degradation}
  \mathbf{y} = \mathbf{x}\downarrow^{bicubic}_{sf}+\mathbf{n},
\end{equation}
where $\downarrow^{bicubic}_{sf}$ denotes bicubic downsamling with downscaling factor $sf$.

We can then solve \eqref{eq:HQS_DMIR_1} with the following iterative back-projection (IBP) solution
\begin{equation}\label{eq:sisr_backprojection}
  \mathbf{x}_{0} = \mathbf{z}_{0} - \gamma(\mathbf{y} - \mathbf{z}_{0}\downarrow^{bicubic}_{sf})\uparrow^{bicubic}_{sf},
\end{equation}
where $\uparrow^{bicubic}_{sf}$ denotes bicubic interpolation with upscaling factor ${sf}$, $\gamma$ is the step size. Through experiment, we found that it's better to use $\gamma_t=\frac{\gamma}{1+\rho_t}$ which will decrease with time. To get the solution accurately, we IBP for more than one iteration for each timestep $t$.

For bicubic SR, we can also solve \eqref{eq:HQS_DMIR_1} in closed-form with an approximated bicubic kernels $\mathbf{k}$ \cite{zhang2021plug} %\cite{zhang2020deep}
\begin{equation}\label{eq:sisr}
  \mathbf{x}_{0} = \mathcal{F}^{-1}\left(\frac{1}{\rho_t}\Big(\mathbf{d} - \overline{\mathcal{F}(\mathbf{k})} \odot_s\frac{(\mathcal{F}(\mathbf{k})\mathbf{d})\Downarrow_s }{(\overline{\mathcal{F}(\mathbf{k})}\mathcal{F}(\mathbf{k}))\Downarrow_s +\rho_t}\Big)\right),
\end{equation}
where $\mathbf{d} = \overline{\mathcal{F}(\mathbf{k})}\mathcal{F}(\mathbf{y}\uparrow_{sf}) + \rho_t\mathcal{F}(\mathbf{z}_{0})$ and $\uparrow_{sf}$ denotes the standard s-fold upsampler, and where $\odot_s$ denotes distinct block processing operator with element-wise multiplication, $\Downarrow_s$ denotes distinct block downsampler, i.e., averaging the $s\times s$ distinct blocks.

In general, the closed-form solution~\eqref{eq:sisr} should outperform iterative solutions~\eqref{eq:sisr_backprojection} in quantitative metrics, since the former contains fewer hyperparameters.

% \clearpage 
\section{Additional Ablation Study}
\label{append:AblationStudy}
\begin{figure}
\centering
\begin{overpic}[width=1.\linewidth]{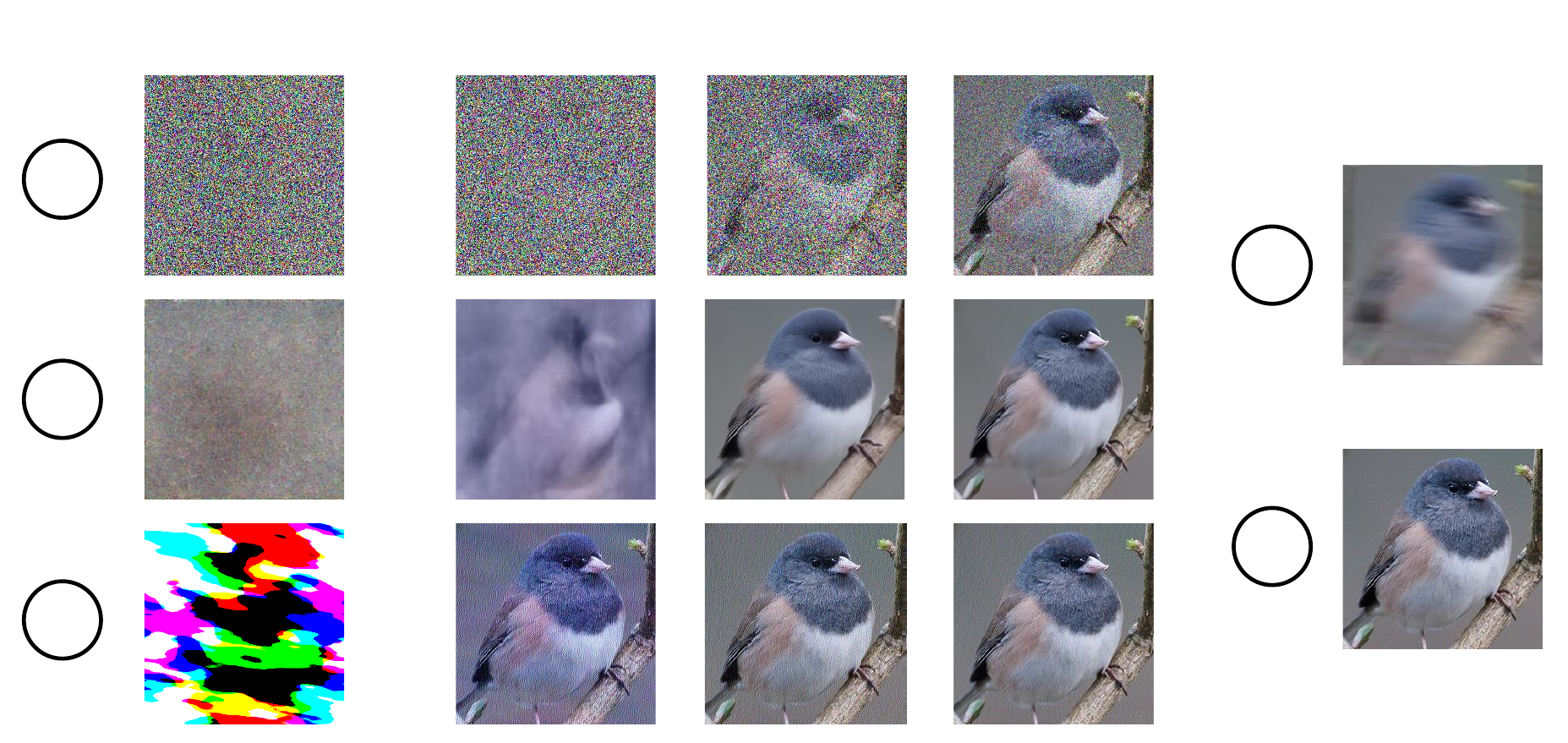}% no need to specify the file extension
\put(76.5, 1.){\color{gray}{\line(0,1){41.4}}}
\put(23.,35.5){\color{black}{$\mathbf{\cdots}$}}
\put(23.,20.5){\color{black}{$\mathbf{\cdots}$}}
\put(23.,6.5){\color{black}{$\mathbf{\cdots}$}}
\put(8.,43.5){\color{black}{\small $t=1000$}}
\put(29,43.5){\color{black}{\small $t=600$}}
\put(45,43.5){\color{black}{\small $t=300$}}
\put(60.4,43.5){\color{black}{\small $t=100$}}
\put(2.2,35.3){\color{black}{\scalebox{0.8}{$\mathbf{x}_t$}}}
\put(2.,20.8){\color{black}{\scalebox{0.8}{$\mathbf{x}^{\scalebox{0.5}{(t)}}_0$}}}
\put(1.1,7){\color{black}{\scalebox{0.8}{ $\mathbf{\hat{x}}^{\scalebox{0.5}{(t)}}_0$}}}
\put(80.0,29.8){\color{black}{\scalebox{0.8}{$\mathbf{y}$}}}
\put(79.5,11.8){\color{black}{\scalebox{0.8}{$\mathbf{x}_0$}}}
\end{overpic}
\caption{Reverse diffusion process in DiffPIR}
\label{fig:ablation_t_start_append}
\vspace{0.2cm}
\end{figure}
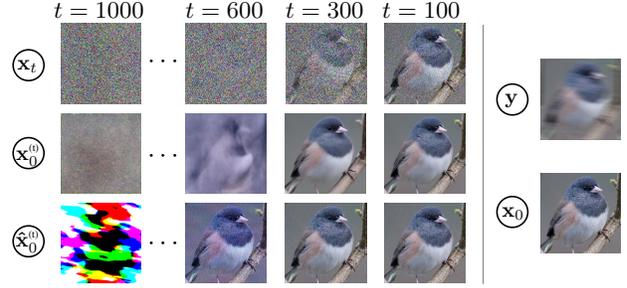

In this section, we illustrate the reverse diffusion process by showing the intermediate results in Figure \ref{fig:ablation_t_start_append}. We observed that in the beginning, the analytical solution offers no help and motivate us to skip this phase. As mentioned in Section \ref{sec:ablation}, we found by experiment $t_{start}$ the end timestep for this phase.

% \clearpage 
\section{Additional Visual Results}
\label{append:VisualResults}
\begin{figure*}
\centering
\begin{overpic}[width=0.85\linewidth]{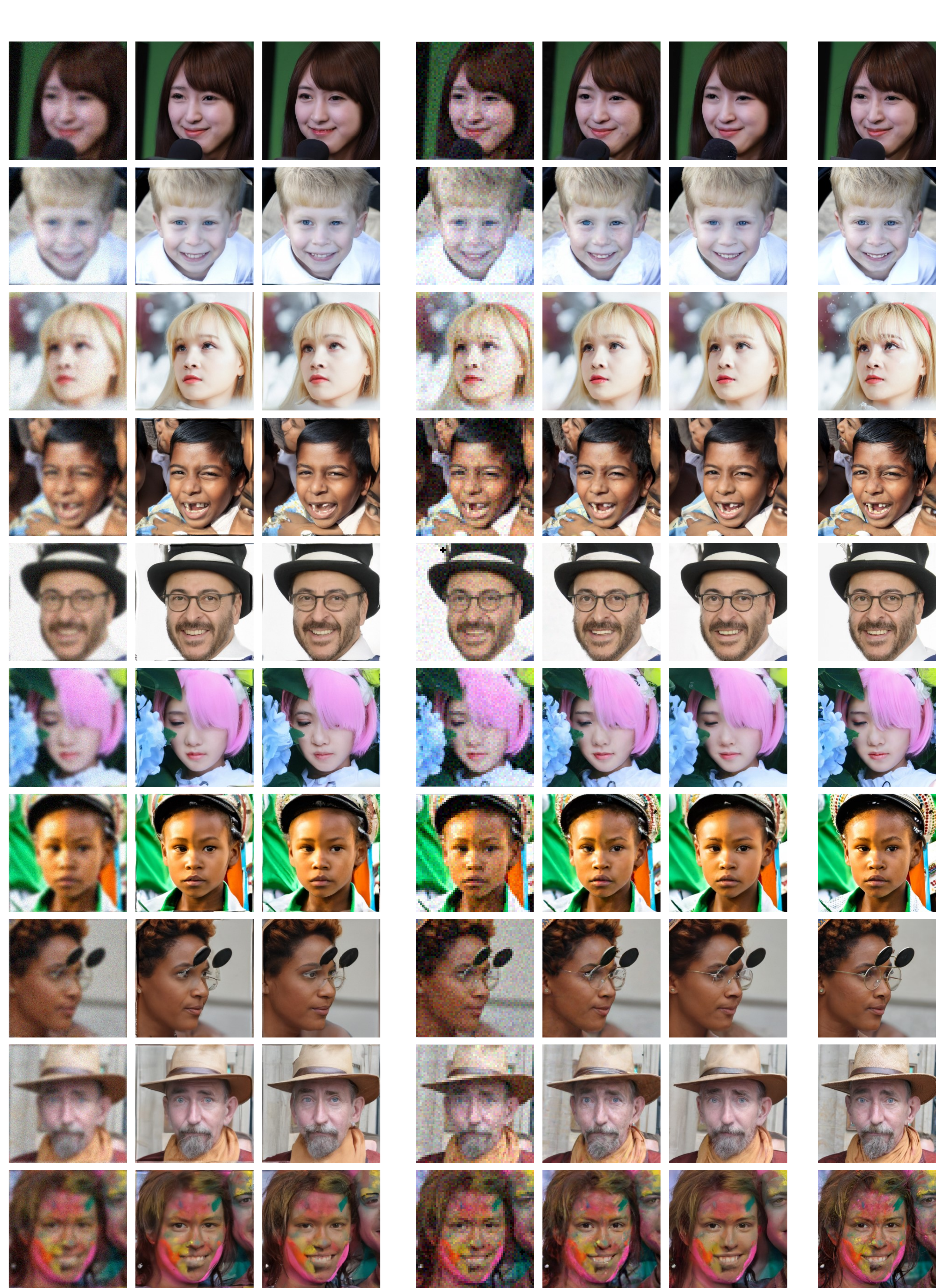}% no need to specify the file extension
%\begin{sideways}
\put(1.,97.2){\color{black}{\small Measurement}}
\put(13,97.2){\color{black}{\small DPS$y_t$}}
\put(23,97.2){\color{black}{\small DPS$y_0$}}
\put(32.5,97.2){\color{black}{\small Measurement}}
\put(44.5,97.2){\color{black}{\small DPS$y_t$}}
\put(54.5,97.2){\color{black}{\small DPS$y_0$}}
\put(63.8,97.2){\color{black}{\small Ground Truth}}
%\end{sideways}
\end{overpic}
\caption{Qualitative results of DPS$y_t$ and DPS$y_0$ (both 1000 NFEs) for Gaussian deblurring (left) and 4$\times$ SR (right) with $\sigma_n=0.05$}
\label{fig:figures_supplementary3}
\vspace{-0.1cm}
\end{figure*}

\begin{figure*}
\centering
\begin{overpic}[width=0.95\linewidth]{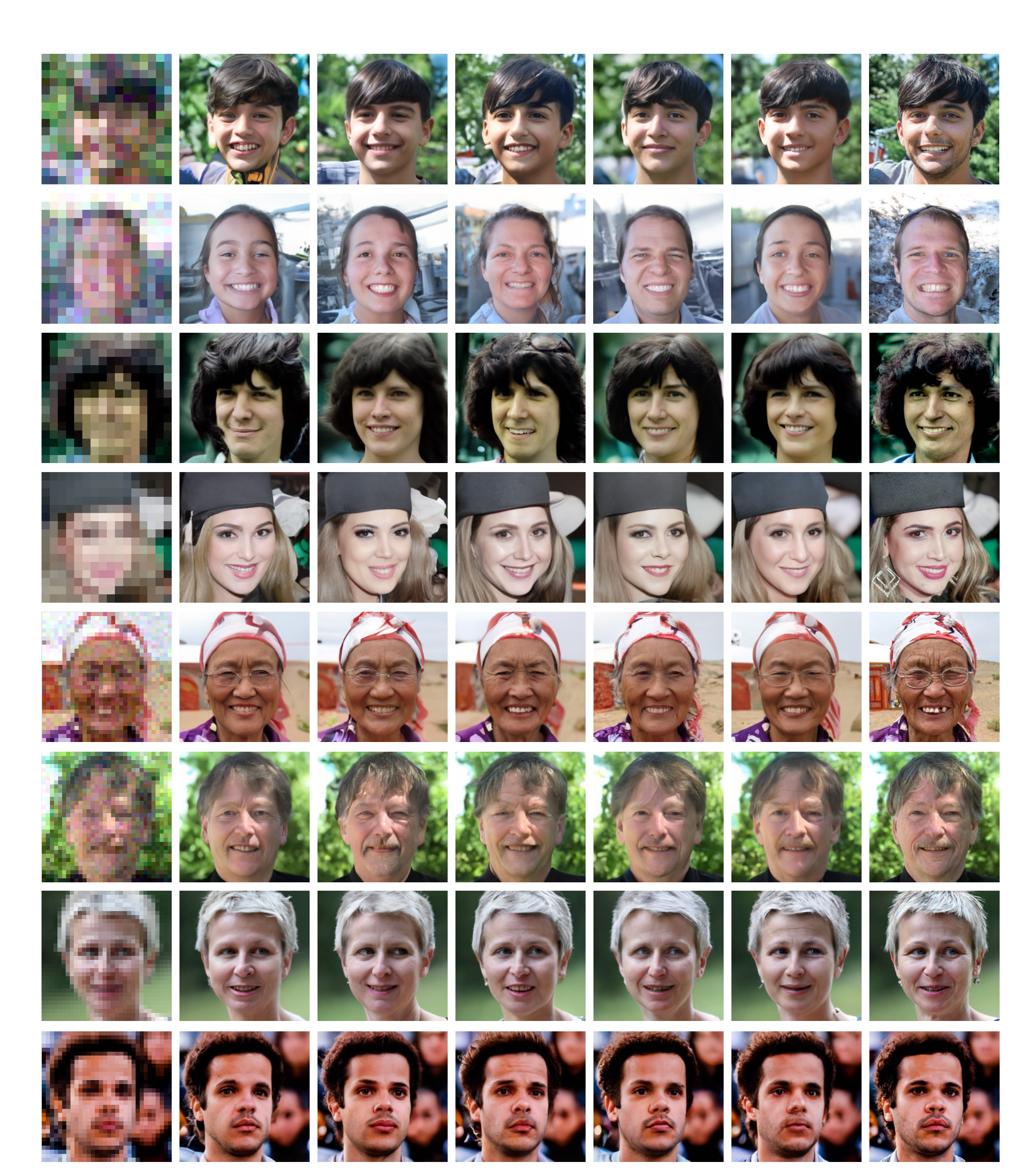}% no need to specify the file extension
\put(4.5,96){\color{black}{\small Measurement}}
\put(17.8,96){\color{black}{\small Sample 1}}
\put(29.3,96){\color{black}{\small Sample 2}}
\put(41.2,96){\color{black}{\small Sample 3}}
\put(52.5,96){\color{black}{\small Sample 4}}
\put(64.5,96){\color{black}{\small Sample 5}}
\put(74.8,96){\color{black}{\small Ground Truth}}
\begin{turn}{90}
\put(7.5,-2){\color{black}{\small 8$\times$ SR ($\sigma_n=0.0$)}}
\put(30,-2){\color{black}{\small 8$\times$ SR ($\sigma_n=0.05$)}}
\put(54.2,-2){\color{black}{\small 16$\times$ SR ($\sigma_n=0.0$)}}
\put(77.2,-2){\color{black}{\small 16$\times$ SR ($\sigma_n=0.05$)}}
\end{turn}
\end{overpic}
\caption{Qualitative results of DiffPIR (100 NFEs) for 8$\times$ and 16$\times$ SR with $\sigma_n=0.0$ and $\sigma_n=0.05$.}
\label{fig:figures_supplementary2}
\vspace{-0.1cm}
\end{figure*}

\begin{figure*}
\centering
\begin{overpic}[width=0.99\linewidth]{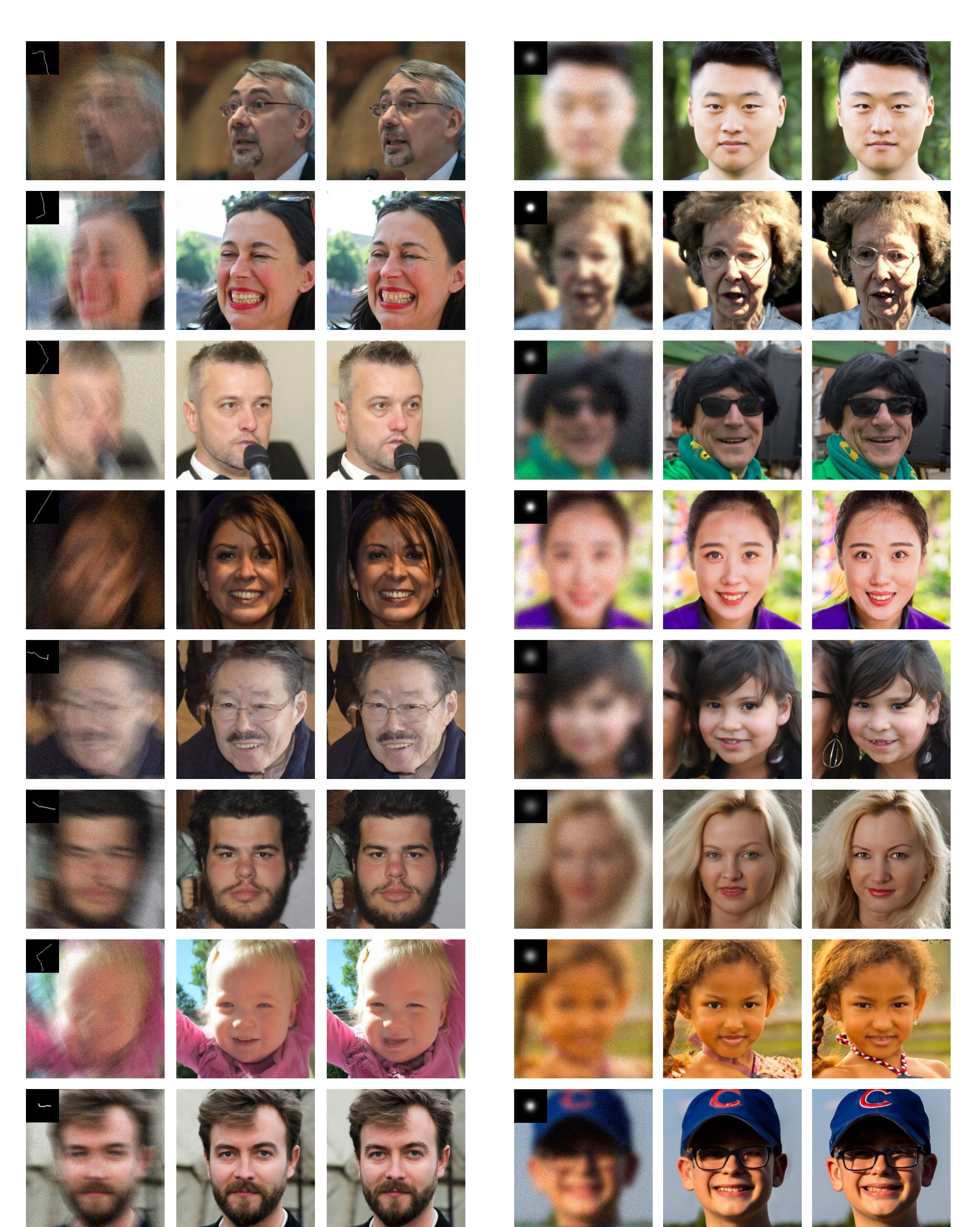}% no need to specify the file extension
%\begin{sideways}
\put(3.7,97.2){\color{black}{\small Measurement}}
\put(17.3,97.2){\color{black}{\small DiffPIR}}
\put(27.8,97.2){\color{black}{\small Ground Truth}}
\put(43.4,97.2){\color{black}{\small Measurement}}
\put(56.9,97.2){\color{black}{\small DiffPIR}}
\put(67.1,97.2){\color{black}{\small Ground Truth}}
%\end{sideways}
\end{overpic}
\caption{Qualitative results of DiffPIR (100 NFEs) for motion deblurring (left) and Gaussian deblurring (right) with $\sigma_n=0.05$}
\label{fig:figures_supplementary1}
\vspace{-0.1cm}
\end{figure*}

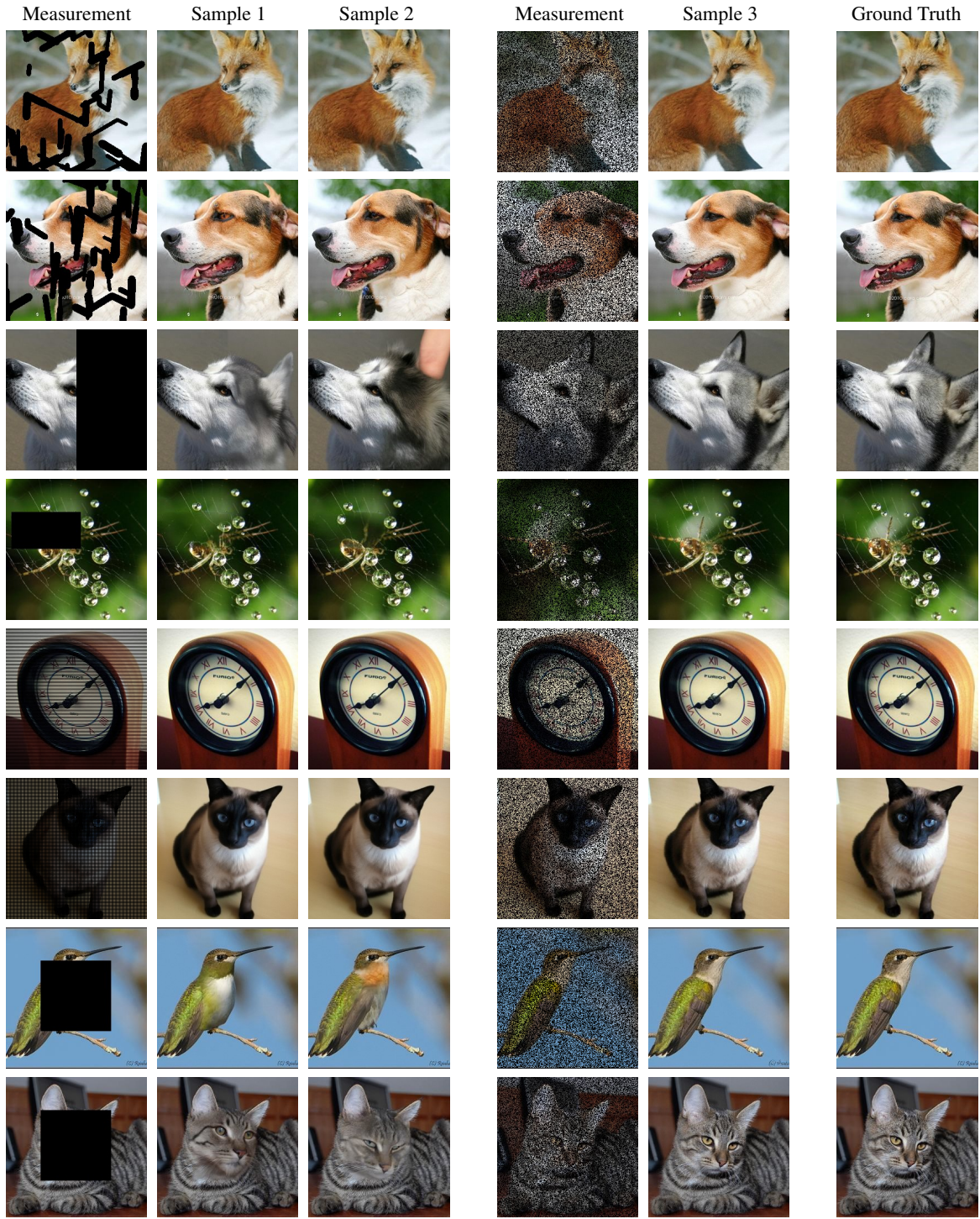
\begin{figure*}
\centering
\begin{overpic}[width=0.90\linewidth]{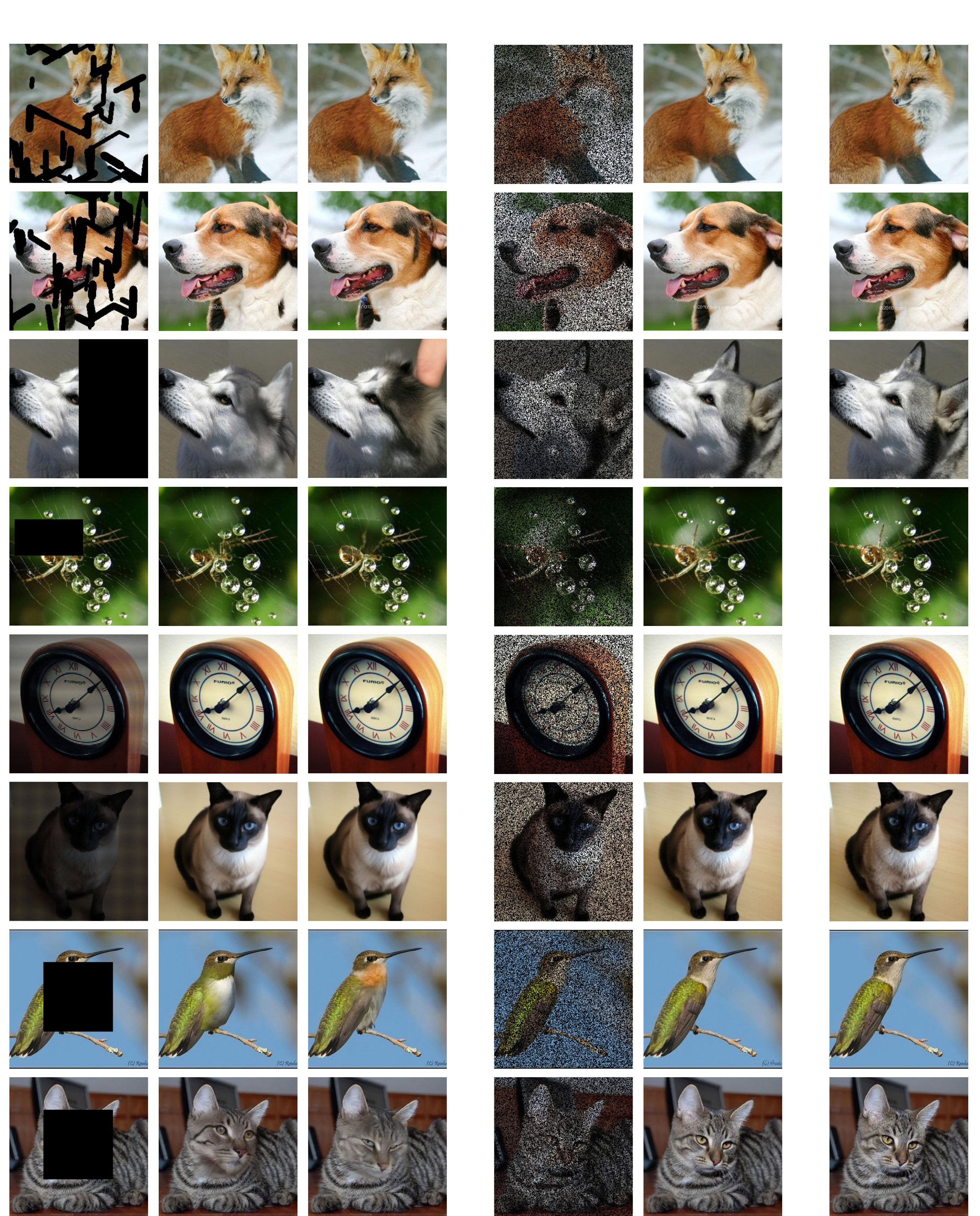}% no need to specify the file extension
%\begin{sideways}
\put(2.1,97.2){\color{black}{\small Measurement}}
\put(15.8,97.2){\color{black}{\small Sample 1}}
\put(27.8,97.2){\color{black}{\small Sample 2}}
\put(42.1,97.2){\color{black}{\small Measurement}}
\put(55.7,97.2){\color{black}{\small Sample 3}}
\put(69.4,97.2){\color{black}{\small Ground Truth}}
%\end{sideways}
\end{overpic}
\caption{Qualitative results of DiffPIR (100 NFEs) for inpainting with different masks ($\sigma_n=0.0$)}
\label{fig:figures_supplementary0}
\vspace{-0.1cm}
\end{figure*}

In this section, we provide additional visual examples for FFHQ and ImageNet datasets to show the ability of our method. 
In Figure \ref{fig:figures_supplementary3} we demonstrate that DPS$y_t$ and DPS$y_0$ both work well on IR tasks like deblurring and SR. In Figure \ref{fig:figures_supplementary2}, we show the diversity of SR reconstructions with diffusion model as generative prior. In Figure \ref{fig:figures_supplementary1} and \ref{fig:figures_supplementary0}, we show that our proposed DiffPIR is capable to handle various blur kernels (both motion and Gaussian) and masks, respectively.

\end{document}